\documentclass[journal,10pt]{IEEEtran}

\usepackage{cite}
\usepackage{amsmath,amssymb,amsfonts}
\usepackage{graphicx}
\usepackage{textcomp}
\usepackage[table]{xcolor}
\usepackage{colortbl}
\usepackage{booktabs}
\usepackage{multirow}
\usepackage{enumitem}
\usepackage{array}
\usepackage{placeins}
\usepackage{caption}
\usepackage{subcaption}
\usepackage{url}
\usepackage{microtype}
\usepackage{tikz}
\usetikzlibrary{positioning,arrows.meta,calc,fit,backgrounds,shapes.geometric,decorations.pathreplacing,shadows.blur,shadows}
\usepackage{cleveref}
\usepackage{etoolbox}

\DeclareRobustCommand{\trainedicon}{%
  \tikz[baseline=-0.4ex, line cap=round, line join=round]{%
    \fill[red!75!orange]
      (0,-0.20)
      .. controls (0.20,-0.05) and (0.13,0.08) ..  (0.05,0.20)
      .. controls (0.10,0.08) and (-0.04,0.04) .. (-0.06,-0.04)
      .. controls (-0.18,0.05) and (-0.18,-0.10) .. (-0.10,-0.16)
      .. controls (-0.05,-0.20) and (-0.02,-0.21) .. (0,-0.20) -- cycle;
    \fill[yellow!90]
      (0,-0.12)
      .. controls (0.10,-0.04) and (0.05,0.08) .. (0,0.12)
      .. controls (-0.04,0.04) and (-0.06,-0.04) .. (-0.04,-0.10)
      .. controls (-0.02,-0.13) and (-0.01,-0.13) .. (0,-0.12) -- cycle;
  }%
}
\DeclareRobustCommand{\frozenicon}{%
  \tikz[baseline=-0.4ex, line cap=round, line join=round, line width=0.7pt]{%
    \pgfmathsetmacro{\R}{0.20}%
    \foreach \a in {0,60,120,180,240,300} {
      \draw[blue!60!cyan] (0,0) -- ({\R*cos(\a)},{\R*sin(\a)});
      \draw[blue!60!cyan] ({0.6*\R*cos(\a)},{0.6*\R*sin(\a)})
        -- ({0.6*\R*cos(\a)+0.06*cos(\a+50)},{0.6*\R*sin(\a)+0.06*sin(\a+50)});
      \draw[blue!60!cyan] ({0.6*\R*cos(\a)},{0.6*\R*sin(\a)})
        -- ({0.6*\R*cos(\a)+0.06*cos(\a-50)},{0.6*\R*sin(\a)+0.06*sin(\a-50)});
    }
    \fill[blue!60!cyan] (0,0) circle (0.025);
  }%
}

\definecolor{ieeeblue}{rgb}{0.21,0.49,0.74}
\colorlet{cvprblue}{ieeeblue}
\colorlet{novelty}{red!75!orange}

\crefname{section}{Sec.}{Secs.}
\crefname{table}{Table}{Tables}
\crefname{figure}{Fig.}{Figs.}
\crefname{equation}{Eq.}{Eqs.}
\crefname{paragraph}{Sec.}{Secs.}

\begin{document}

\title{Adjudicated Captioning: Multi-Agent Alignment Scoring and\\
Consensus-Distilled Beam Arbitration for\\
Strict Zero-Shot Image Captioning}

\author{%
  Duy~Tran~Thanh\textsuperscript{*},
  Thien-Phuc~Doan,
  Long~Nguyen-Vu,
  Ngo~Tan~Vu~Khanh%
  \thanks{\textsuperscript{*}Corresponding author. Duy Tran Thanh is a Senior
  AI Engineer with AI Platform OneNexus, OneMount
  (e-mail: duy.tran10@onemount.com).}%
  \thanks{Thien-Phuc Doan is with the School of Electronic Engineering, Soongsil
  University, Seoul 06978, South Korea (e-mail: phucdt@ssu.ac.kr).}%
  \thanks{Long Nguyen-Vu is with the AI Laboratory, MoAdata, Seoul, South Korea
  (e-mail: longnv@moadata.ai).}%
  \thanks{Ngo Tan Vu Khanh is with the University of Economics Ho Chi Minh City
  (UEH), Ho Chi Minh City, Vietnam (e-mail: khanhntv@ueh.edu.vn).}%
}

\maketitle
\pagestyle{plain}
\thispagestyle{plain}

\begin{abstract}
Zero-shot image captioning aims to describe images in natural language without paired image--caption supervision during Captioner training, relying instead on text-only corpora and frozen pretrained image--text scorers. Existing retrieval-augmented methods score image--text alignment exactly once, at the initial retrieval step, and then commit the Captioner's autoregressive beam under language-model probability alone, leaving the decoder without further visual grounding feedback. Progress in this regime has consequently stalled, with no method improving on the strict-regime best since $2024$. We propose \textbf{Adjudicated Captioning}, an inference-time multi-agent framework that restores grounding feedback at multiple checkpoints over an unchanged IFCap Captioner and crosses the plateau without paired image--caption supervision. Our contributions are threefold. First, we install a stronger frozen Retrieval Encoder at the input. Second, between retrieval and decoding we insert a frozen Cross-Attention Verifier that re-ranks the top-$9$ retrievals to top-$5$, producing a Verifier-adjudicated retrieval pool. Third, at the output beam we attach a learned Reranker pairing \textbf{TriFuse}, a multilayer perceptron, with \textbf{MemAttend}, a memory-attended transformer, the pipeline's only learned components; both are trained self-supervised by Borda-consensus distillation across the three frozen scorers, consuming no paired image--caption labels and no reference captions. To our knowledge these are the first learned Beam Rerankers proposed under the captioner-text-only regime. Under the inductive headline protocol, with Rerankers fit on the disjoint COCO Karpathy validation beam and applied frozen to test, the framework reaches \textbf{CIDEr} (Consensus-based Image Description Evaluation) $\mathbf{117.6}$ and \textbf{SPICE} (Semantic Propositional Image Caption Evaluation) $\mathbf{21.9}$ on COCO Karpathy, up from $108.0$ and $20.3$ for IFCap, a $+9.6$ CIDEr gain, and $+7.7$ above Negative Entity Suppression (NES), the strongest synthetic-image-augmented method at $109.9$, without retraining the Captioner. A training-free fixed-fusion baseline reaches $115.8$ CIDEr, so $+7.8$ of the $+9.6$ gain is attributable to the non-learned architectural intervention at the retrieval and verification checkpoints, and the remaining $+1.8$ to the learned Rerankers. The same recipe transfers off-COCO without Captioner retraining: $+8.1$ CIDEr on Flickr30k Karpathy and $+5.7$ on NoCaps overall. A Verifier-substrate swap to a non-COCO-fine-tuned matcher still yields $114.0$ CIDEr, $+6.0$ over IFCap, showing the gain is not solely conditional on COCO-specific Verifier tuning.
\end{abstract}


\IEEEpeerreviewmaketitle

\definecolor{novelty}{RGB}{220,68,46}
\definecolor{noveltyfill}{RGB}{255,236,228}
\definecolor{noveltyedge}{RGB}{198,58,38}
\definecolor{stageink}{RGB}{36,46,68}
\definecolor{stagefill}{RGB}{246,248,252}
\definecolor{stageedge}{RGB}{122,138,166}
\definecolor{coreink}{RGB}{72,72,80}
\definecolor{priorgray}{RGB}{128,128,134}
\definecolor{priorfill}{RGB}{240,240,244}
\definecolor{priorink}{RGB}{90,90,100}
\definecolor{captionbubble}{RGB}{255,250,238}
\definecolor{captionedge}{RGB}{204,150,52}
\definecolor{checkblue}{RGB}{60,103,179}
\definecolor{checkblueFill}{RGB}{230,238,250}

\newsavebox{\teaserbox}
\savebox{\teaserbox}{%
\begin{tikzpicture}[
  font=\sffamily\small,
  >={Latex[length=2.2mm,width=1.7mm]},
  priorcard/.style={draw, line width=0.6pt, rounded corners=2.5pt,
                    fill=priorfill, draw=priorgray, text=priorink,
                    align=center, font=\sffamily\fontsize{7.5}{9}\selectfont,
                    inner sep=1.7mm, minimum height=12mm,
                    minimum width=26mm},
  ourcard/.style={draw, line width=0.7pt, rounded corners=2.5pt,
                  fill=stagefill, draw=stageedge, text=stageink,
                  align=center, font=\sffamily\fontsize{7.5}{9}\selectfont,
                  inner sep=1.7mm, minimum height=12mm,
                  minimum width=26mm,
                  drop shadow={shadow xshift=0.4mm, shadow yshift=-0.4mm,
                               opacity=0.18, fill=black!60}},
  ournovel/.style={ourcard, fill=noveltyfill, draw=noveltyedge,
                   text=noveltyedge!80!black, line width=1.0pt},
  checkmark/.style={circle, draw, line width=0.7pt, fill=checkblueFill,
                    draw=checkblue, text=checkblue,
                    font=\sffamily\bfseries\fontsize{6.5}{7.5}\selectfont,
                    inner sep=0pt, minimum size=4.8mm},
  checkmarknov/.style={checkmark, fill=noveltyfill, draw=noveltyedge,
                       text=noveltyedge},
  priorarr/.style={->, line width=0.7pt, draw=priorgray!130!black, >=Latex},
  ourarr/.style={->, line width=1.1pt, draw=novelty!85!black, >=Latex},
  imgframe/.style={draw=black!50, line width=0.5pt, rounded corners=1pt,
                   inner sep=0pt,
                   drop shadow={shadow xshift=0.4mm, shadow yshift=-0.4mm,
                                opacity=0.18, fill=black!60}},
  imgframeprior/.style={draw=priorgray!80!black, line width=0.5pt,
                        rounded corners=1pt, inner sep=0pt},
  capprior/.style={draw=priorgray, line width=0.5pt, rounded corners=2.5pt,
                   fill=priorfill, text width=36mm, align=left, text=priorink,
                   inner sep=1.6mm, font=\sffamily\fontsize{7}{8.4}\selectfont\itshape},
  capour/.style={draw=captionedge, line width=0.8pt, rounded corners=2.5pt,
                 fill=captionbubble, text width=36mm, align=left,
                 inner sep=1.6mm, font=\sffamily\fontsize{7}{8.4}\selectfont\itshape,
                 drop shadow={shadow xshift=0.4mm, shadow yshift=-0.4mm,
                              opacity=0.18, fill=black!60}},
  scoreprior/.style={draw=priorgray, line width=0.5pt, rounded corners=2pt,
                     fill=priorfill, text=priorink,
                     font=\sffamily\bfseries\fontsize{8}{9.6}\selectfont,
                     inner sep=1.6mm, align=center, minimum width=20mm},
  scoreour/.style={draw=noveltyedge, line width=1.0pt, rounded corners=2pt,
                   fill=noveltyfill, text=noveltyedge!85!black,
                   font=\sffamily\bfseries\fontsize{8}{9.6}\selectfont,
                   inner sep=1.6mm, align=center, minimum width=20mm,
                   drop shadow={shadow xshift=0.4mm, shadow yshift=-0.4mm,
                                opacity=0.18, fill=black!60}},
  rowtag/.style={font=\sffamily\bfseries\fontsize{8.5}{10}\selectfont,
                 anchor=east, align=right},
  rowtaglight/.style={rowtag, text=priorink},
  rowtagnov/.style={rowtag, text=noveltyedge!85!black},
  newbadge/.style={fill=novelty, text=white,
                   font=\sffamily\bfseries\fontsize{6}{7}\selectfont,
                   rounded corners=1.2pt, inner xsep=2.4pt, inner ysep=1.2pt},
  annot/.style={font=\sffamily\fontsize{6.5}{7.5}\selectfont\itshape,
                text=coreink!75},
  contribution/.style={font=\sffamily\bfseries\fontsize{7}{8.4}\selectfont,
                       text=noveltyedge!85!black, fill=noveltyfill,
                       inner sep=1.0mm, rounded corners=1.5pt,
                       draw=noveltyedge, line width=0.6pt},
]

\node[imgframeprior] (p_img) at (0,0)
  {\includegraphics[width=12mm, height=12mm, keepaspectratio]
     {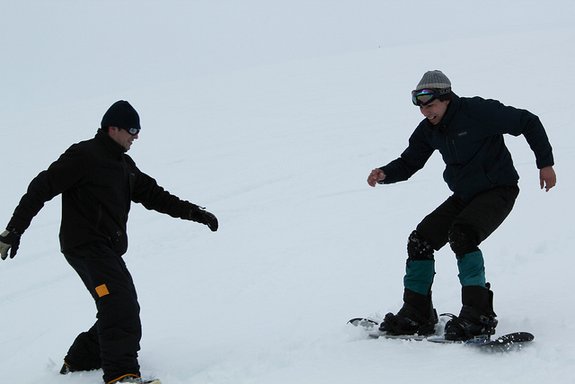}};

\node[priorcard, right=8mm of p_img] (p_ret) {Retrieval\,\frozenicon\\\fontsize{6.5}{7.5}\selectfont\itshape ViT-B/32};
\node[checkmark, above=2mm of p_ret] {\checkmark};

\node[priorcard, right=8mm of p_ret] (p_cap) {Captioner\,\frozenicon};

\node[capprior, right=8mm of p_cap] (p_out)
  {``A snowboarder and a snowboarder are posing\dots''};

\node[scoreprior, right=4mm of p_out] (p_score) {CIDEr $108.0$};

\draw[priorarr] (p_img.east)  -- (p_ret.west);
\draw[priorarr] (p_ret.east)  -- (p_cap.west);
\draw[priorarr] (p_cap.east)  -- (p_out.west);

\node[rowtaglight] at ($(p_img.west)+(-2.5mm,0)$) {Prior};

\draw[draw=black!18, line width=0.4pt, dash pattern=on 1.5pt off 1.6pt]
  ($(p_img.south west)+(-30mm,-5mm)$) -- ($(p_score.south east)+(2mm,-5mm)$);

\node[imgframe, below=18mm of p_img.south, anchor=north] (o_img)
  {\includegraphics[width=15mm, height=15mm, keepaspectratio]
     {figures/qual_data/qual_coco/imgs/COCO_val2014_000000127451.jpg}};

\node[ournovel, right=8mm of o_img] (o_ret)
  {Retrieval\,\frozenicon\\\fontsize{6.5}{7.5}\selectfont\itshape ViT-bigG/14};
\node[checkmarknov, above=2mm of o_ret] {\checkmark};
\node[newbadge, below=0.6mm of o_ret] {NEW};

\node[ournovel, right=7mm of o_ret] (o_ver)
  {Verifier\,\frozenicon\\\fontsize{6.5}{7.5}\selectfont\itshape BLIP-ITM cross-attn};
\node[checkmarknov, above=2mm of o_ver] {\checkmark};
\node[newbadge, below=0.6mm of o_ver] {NEW};

\node[ourcard, right=7mm of o_ver] (o_cap)
  {Captioner\,\frozenicon};

\node[ournovel, right=7mm of o_cap] (o_rer)
  {Learned Reranker\,\trainedicon\\\fontsize{6.5}{7.5}\selectfont\itshape TriFuse $+$ MemAttend};
\node[checkmarknov, above=2mm of o_rer] {\checkmark};
\node[newbadge, below=0.6mm of o_rer] {NEW};

\node[capour, right=7mm of o_rer] (o_out)
  {``Two men are snowboarding down a snowy hill.''};

\node[scoreour, right=4mm of o_out] (o_score)
  {CIDEr $\mathbf{117.6}$\\[-0.2mm]\fontsize{6.5}{7.5}\selectfont\color{novelty}$+9.6$};

\draw[ourarr] (o_img.east)  -- (o_ret.west);
\draw[ourarr] (o_ret.east)  -- (o_ver.west);
\draw[ourarr] (o_ver.east)  -- (o_cap.west);
\draw[ourarr] (o_cap.east)  -- (o_rer.west);
\draw[ourarr] (o_rer.east)  -- (o_out.west);

\node[rowtagnov] at ($(o_img.west)+(-2.5mm,0)$) {Ours};

\end{tikzpicture}%
}

\begin{figure*}[t!]
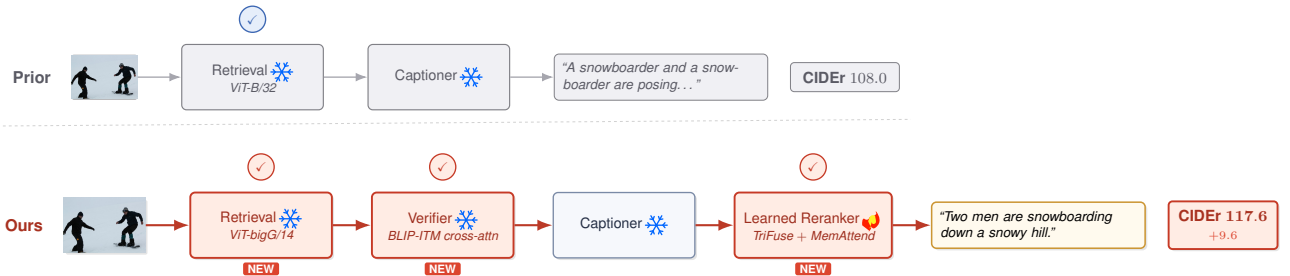

\centering
\resizebox{\textwidth}{!}{\usebox{\teaserbox}}
\caption{\textbf{From single-checkpoint to multi-checkpoint alignment scoring.} Prior strict-ZIC pipelines (top) score image--text alignment at exactly one place: the retrieval-time cosine ({\color{checkblue}\checkmark}). Our pipeline (bottom) adds three novel alignment-scoring components ({\color{novelty}\textbf{NEW \checkmark}}): a stronger Retrieval encoder, a Cross-Attention Verifier, and a learned Beam Reranker. The Captioner is left untouched. The three new heads are trained self-supervised over the three frozen scorers, with no paired image--caption labels. Result: $+9.6$ CIDEr on COCO Karpathy ($117.6$ vs.\ $108.0$).}
\label{fig:teaser}
\end{figure*}

\section{Introduction}
\label{sec:intro}

Retrieval-augmented image captioning has converged on a single design rule: image--text alignment is scored exactly once, at retrieval, and is not consulted again. ClipCap~\cite{mokady2021clipcap} fixed the recipe. A small mapping network turns a CLIP image embedding into a soft prefix for a frozen GPT-2~\cite{radford2019gpt2}; under the strict text-only zero-shot image captioning (ZIC) regime the mapping network sees no image during training and bridges the modality gap at inference using CLIP cosines. Subsequent work, ranging from CapDec~\cite{nukrai2022capdec} and DeCap~\cite{li2023decap} to ViECap~\cite{fei2023viecap}, MeaCap~\cite{zeng2024meacap}, and IFCap~\cite{lee2024ifcap}, has refined this recipe in different places, but every method commits the Captioner's beam under language-model probability after retrieval, without any further image--text alignment check.

Progress under this rule has been uneven, and it has stalled outright since $2024$. On COCO Karpathy~\cite{lin2014coco,karpathy2015deepvs}, CIDEr\footnote{Metric acronyms used throughout: CIDEr, Consensus-based Image Description Evaluation; SPICE, Semantic Propositional Image Caption Evaluation; BLEU, Bilingual Evaluation Understudy; METEOR, Metric for Evaluation of Translation with Explicit ORdering; ROUGE-L, Recall-Oriented Understudy for Gisting Evaluation, longest-common-subsequence variant.} improved from $92.9$ with ViECap to $95.4$ with MeaCap, a $+2.5$ step, and then to $108.0$ with IFCap, a larger $+12.6$ step delivered by two simultaneous changes, image-like retrieval and frequency-based entity filtering, rather than by any single new component. Since IFCap, no method operating inside the strict regime has, to our knowledge, improved on $108.0$. The one entry above it, Negative Entity Suppression (NES)~\cite{lu2026negative} at $109.9$, gains $+1.9$ and does so only by departing from the regime and training on Stable-Diffusion-generated images alongside the text corpus. Successive strict-ZIC methods have each introduced a new \emph{filter} between retrieval and decoding, including entity-aware masking in ViECap, frequency-based entity filtering in IFCap, and attention-level suppression in NES; every one of these is a further constraint on a single retrieval step rather than a new alignment-scoring event.

We attribute the plateau to a structural property of the pipeline that has remained unchallenged since retrieval-augmented captioning was introduced, rather than to a limit of language modelling. Image--text alignment is consulted exactly once, at the initial CLIP-cosine retrieval, after which the decoder operates on its own. The Captioner's beam is selected by language-model probability and is therefore biased toward fluent outputs that need not be fully grounded; the retrieval pool itself remains noisy because a dual-encoder contrastive model orders out-of-distribution images only approximately. No prior strict-ZIC system inserts a stronger image--text scorer at any later point in the pipeline.

Building on this observation, we propose a strict-ZIC framework that scores image--text alignment at \emph{multiple} alignment checkpoints rather than only at retrieval, as illustrated in Figure~\ref{fig:pipeline}. Concretely, we make three architectural changes on top of an unchanged Captioner. First, we replace IFCap's OpenAI CLIP ViT-B/32~\cite{radford2021clip} dual encoder with a scaling-law-stronger OpenCLIP ViT-bigG/14~\cite{cherti2022openclip} for top-$9$ retrieval against the $566$K-caption COCO Karpathy training corpus, which is the largest single-component lift in our ablation. We refer to the two by their architecture names, ViT-B/32 and ViT-bigG/14, throughout. Second, we insert a frozen Cross-Attention Verifier between retrieval and decoding to re-rank the top-$9$ down to top-$5$; instantiated with the BLIP image--text matching head~\cite{li2022blip}, the Verifier's rank correlation with the Retrieval Encoder cosine is only $\rho \!\approx\! 0.4$ on IFCap retrievals, providing partly orthogonal information that the dual encoder cannot supply. Third, at the Captioner's output we attach two complementary self-supervised learned Beam Rerankers. \textbf{TriFuse} is a $113$-parameter multilayer perceptron (MLP) that fuses three frozen-scorer signals per beam candidate: the GPT-2 log-probability, the Retrieval Encoder cosine, and the Verifier's matching score. \textbf{MemAttend} is a $17.4$K-parameter memory-augmented transformer encoder that lets candidates cross-attend to the Verifier-adjudicated retrieval pool. Both Rerankers are trained by listwise Borda-consensus distillation across the three frozen scorers; no ground-truth captions and no paired image--caption supervision are used at any point, so the strict text-only training regime is preserved end-to-end. The Captioner itself, namely IFCap's released mapping network and GPT-2 decoder, is never retrained or fine-tuned.

We adopt the operational definition used by every prior strict-ZIC method: the regime constrains \emph{Captioner training}, not the externally pretrained frozen scorers consumed at inference (\Cref{sec:related_regime}). To avoid ambiguity we use the more precise label \textbf{captioner-text-only ZIC under frozen paired-pretrained scorers} where it matters. Within this regime, our \emph{inductive headline}, with TriFuse and MemAttend trained on the disjoint COCO Karpathy \emph{validation} beam-$20$ dump and applied frozen to the test split, reaches CIDEr $\mathbf{117.6}$ / SPICE $\mathbf{21.9}$ / BLEU-$4$ $\mathbf{32.8}$ / METEOR $\mathbf{27.9}$ / ROUGE-L $\mathbf{55.5}$, $+9.6$ CIDEr over the previous strict-ZIC state of the art, IFCap, and $+7.7$ over the synthetic-image-augmented NES. A transductive variant, trained on the test images' frozen-scorer signals (no reference captions used), differs by at most $0.1$ on any primary metric, and not at all on CIDEr, and is reported as a strict upper bound, not the headline. A training-free fixed-fusion baseline reaches $115.8$ CIDEr without any learned head, isolating the architectural intervention from the learned step. The same inference-only recipe transfers off-COCO without Captioner retraining: $+8.1$ CIDEr on Flickr30k Karpathy~\cite{young2014flickr30k,plummer2017flickr30k} and $+5.7$ on NoCaps~\cite{agrawal2019nocaps} overall, with consistent in-/near-/out-domain gains (\Cref{tab:xdomain}, benchmark-transductive). A reference-free CLIPScore (no ground-truth captions consumed) and a reference-proxy Caption Hallucination Assessment with Image Relevance (CHAIR) check corroborate the gain, the latter reducing sentence-level reference-proxy hallucination from $13.28\%$ to $7.84\%$, roughly $41\%$ relative.

In summary, our contributions are as follows:
\begin{itemize}[leftmargin=1.5em,itemsep=0.15em]
  \item We articulate and empirically validate an \emph{architectural} hypothesis for the strict-ZIC plateau, namely that image--text alignment is scored at a single alignment checkpoint, and propose a remedy: place strong, frozen image--text scorers at \emph{multiple} alignment checkpoints, including a Cross-Attention Verifier between retrieval and decoding and a learned Reranker at the Captioner's output.
  \item We introduce two complementary, fully self-supervised Beam Rerankers: \textbf{TriFuse}, a $113$-parameter MLP, and \textbf{MemAttend}, a $17.4$K-parameter memory-augmented transformer encoder. Both are trained by Borda-consensus distillation across three frozen scorers, with no paired image--caption supervision. To our knowledge these are the first learned Beam Rerankers proposed for the captioner-text-only ZIC regime; supervised captioning has used learned beam rerankers (e.g.\ MBR-style decoding~\cite{eikema2022mbr}, ListNet rank objectives~\cite{cao2007listnet}), but those use paired image--caption labels that this regime forbids.
  \item Under the captioner-text-only ZIC regime with frozen paired-pretrained scorers, our pipeline reaches CIDEr $\mathbf{117.6}$ and SPICE $\mathbf{21.9}$ on COCO Karpathy (inductive headline), surpassing IFCap by $+9.6$ and NES by $+7.7$ CIDEr without retraining or fine-tuning the Captioner. The same inference-only recipe transfers off-COCO: $+8.1$ CIDEr on Flickr30k Karpathy and $+5.7$ CIDEr on NoCaps overall.
\end{itemize}

\section{Related Work}
\label{sec:related}

\subsection{Strict zero-shot image captioning}

Strict-ZIC methods can be sorted by \emph{where} they apply image conditioning. ZeroCap~\cite{tewel2022zerocap} and MAGIC~\cite{su2022magic} apply it during decoding, rescoring partial sequences against a CLIP score at each generation step. CapDec~\cite{nukrai2022capdec} pushes the burden upstream by perturbing text features with isotropic Gaussian noise at training time, so the mapping network sees an image-like distribution at inference. DeCap~\cite{li2023decap} avoids the prefix altogether and decodes directly in CLIP space. ViECap~\cite{fei2023viecap} prepends an entity-aware hard prompt, MeaCap~\cite{zeng2024meacap} prepends a memory bank of retrieved captions, and IFCap~\cite{lee2024ifcap} combines image-like noise injection with a frequency-based entity filter on the retrieval pool. The lineage shares one structural commitment: a CLIP-style scorer at retrieval is the only image-conditioning event, after which the Captioner generates without further alignment checks. IFCap reaches CIDEr $108.0$ on COCO Karpathy, the previous strict-text-only best on this benchmark. Paired-supervised foundation models such as BLIP-2~\cite{li2023blip2}, InstructBLIP~\cite{dai2023instructblip}, and LLaVA~\cite{liu2023llava} train on hundreds of millions of paired images and captions and reach far higher CIDEr, with BLIP-2 at $145.8$; they lie outside the strict-text-only regime, and we report them in a separate block of Table~\ref{tab:scoreboard}. We build atop IFCap's released checkpoint without modifying its weights, and our entire intervention is inference-time.

\subsection{Synthetic-image-augmented zero-shot captioning}

A second line replaces text-only training with synthetic image--caption pairs generated from training captions. SynTIC~\cite{liu2024syntic} renders an image per training caption with Stable Diffusion~\cite{rombach2022sd} and trains on those synthetic pairs, reaching $101.1$ CIDEr. PCM-Net~\cite{zeng2024pcmnet} repairs the rendering quality with a patch-wise cross-modal mixup and reaches $103.8$. NES~\cite{lu2026negative}, the strongest entry at $109.9$, layers attention-level negative-entity suppression on the same synthetic-image retrieval pipeline. These methods pay the price of image generation at training time and modify the Captioner; our work does neither. The pipeline we describe is an inference-time intervention on the released IFCap checkpoint, and outperforms NES by $+7.7$ CIDEr on COCO Karpathy while remaining captioner-text-only with frozen paired-pretrained scorers.

\subsection{Image--text alignment scoring and reranking}

Hessel et al.~\cite{hessel2021clipscore} introduced CLIPScore as a reference-free image--text alignment metric, since used to rerank caption beams in supervised captioning. We use it in its ViT-bigG/14 incarnation~\cite{cherti2022openclip} at the output checkpoint of a strict-ZIC pipeline. An output-checkpoint substrate sweep we ran (CLIP variant held against the rest of the pipeline) confirms monotone scaling: CIDEr rises from $111.7$ with OpenAI CLIP ViT-B/32 to $114.4$ with ViT-bigG/14. The full pipeline, with the stronger Retrieval Encoder, the Verifier, the accompanying entity-filter threshold rescale from $K{=}5$ over $L{=}9$ retrievals to $K{=}3$ over $L{=}5$ that keeps the admission fraction fixed once the Verifier shrinks the pool (\Cref{sec:method_pipeline}), and the learned TriFuse+MemAttend Reranker, reaches $117.6$. Cross-encoder reranking over a dual-encoder candidate set is standard in document retrieval~\cite{nogueira2019passage}; listwise learning-to-rank objectives~\cite{cao2007listnet} and minimum-Bayes-risk decoding~\cite{eikema2022mbr} provide the closest parallels for our beam-arbitration step, and ensemble distillation~\cite{hinton2015distillation} motivates our Borda-consensus training target. The cross-encoder rerank has been applied at training time in image captioning: NES~\cite{lu2026negative} uses synthetic-image retrieval as a second-opinion at training time, and PCM-Net~\cite{zeng2024pcmnet} uses a patch-wise mixup to repair the second-opinion. To our knowledge, no prior strict-ZIC method inserts a frozen cross-attention image--text matcher, specifically the BLIP-ITM head~\cite{li2022blip}, at \emph{inference} time as a second-stage rerank, and no prior method places a separate \emph{learned} rerank stage at the Captioner's \emph{output} beam. We empirically demonstrate that placing scorers at \emph{both} alignment checkpoints crosses the strict-ZIC plateau by $+9.6$ CIDEr without retraining the Captioner.

\subsection{Operational definition of the strict-text-only regime}
\label{sec:related_regime}

The term ``strict zero-shot image captioning'' is sometimes interpreted as ``no paired image--text data anywhere in the system,'' an interpretation that no existing method, ours or prior, satisfies. We follow the operational definition used by the strict-ZIC family of DeCap~\cite{li2023decap}, ViECap~\cite{fei2023viecap}, MeaCap~\cite{zeng2024meacap}, IFCap~\cite{lee2024ifcap}, and NES~\cite{lu2026negative}: the constraint applies to the \emph{Captioner training}, not to externally pretrained frozen scorers. The Captioner sees a text corpus alone; the frozen image--text scorers used at inference, namely CLIP variants and the BLIP-ITM head, were pretrained on paired image--text data and are used as-is at evaluation time. Under this definition, prior strict-ZIC work already uses OpenAI CLIP ViT-B/32, trained on $400$M web image--text pairs, at retrieval, and ClipCap-style mapping is itself defined relative to a frozen pretrained CLIP. Our pipeline uses two such scorers. The first is OpenCLIP ViT-bigG/14~\cite{cherti2022openclip} pretrained on LAION-2B image--text pairs. The second is the BLIP image--text matching head~\cite{li2022blip} fine-tuned on COCO image--text pairs. The ViT-bigG/14 pretraining set does not enumerate which images were paired with which text, but the LAION-$2$B distribution overlaps the COCO domain; the BLIP-ITM head was explicitly fine-tuned on COCO captions. Two clarifications are in order. \emph{First}, no COCO Karpathy training/validation/test \emph{caption} is used as a label during our pipeline; retrieval reads the COCO training caption corpus as a text-only index, exactly as IFCap does. \emph{Second}, BLIP-ITM's explicit COCO fine-tuning is not the same as CLIP's broad web-scale pretraining, and we do not claim equivalence; we retain BLIP-ITM as a frozen scorer because it provides a stronger cross-attention image--text matcher than any dual encoder, and we separately report the Flickr30k-fine-tuned BLIP variant (\Cref{tab:verifier_substrate}) so a reader can judge how much of the COCO gain is conditional on that fine-tuning. It is a stronger image--text matcher than CLIP, and our gain decomposes as ``stronger scorer at retrieval'' plus ``stronger scorer at output'' rather than ``access to COCO supervision elsewhere''. Our COCO Karpathy result should be read as a ZIC result \emph{conditional on access to a COCO-tuned frozen Verifier}; we report Flickr30k Karpathy and NoCaps val with the same frozen Verifier in \Cref{sec:abl_xdomain} to show that the gain persists when the Captioner has never been trained on Flickr30k or NoCaps captions. We report a Verifier-substrate ablation in our Verifier-substrate analysis: swapping the COCO-fine-tuned BLIP-ITM head for the Flickr30k-fine-tuned variant of the same architecture~\cite{li2022blip} still yields a $+6.0$ CIDEr lift over IFCap, confirming that the architectural intervention is not an artefact of the Verifier's COCO exposure.

\section{Method}
\label{sec:method}

\newsavebox{\pipeboxA}
\savebox{\pipeboxA}{%
\begin{tikzpicture}[
  font=\small,
  >={Latex[length=2.0mm,width=1.6mm]},
  imgbox/.style={draw, thick, rounded corners=2pt, minimum width=18mm, minimum height=22mm,
                 fill=black!5, align=center, font=\scriptsize},
  modelbox/.style={draw, line width=0.8pt, rounded corners=3pt,
                   minimum width=34mm, minimum height=22mm, align=center,
                   font=\footnotesize, inner sep=2.5mm,
                   drop shadow={shadow xshift=0.4mm, shadow yshift=-0.4mm, opacity=0.25}},
  encmodel/.style={modelbox, fill=cvprblue!18,  draw=cvprblue!60!black},
  blipmodel/.style={modelbox, fill=orange!22,   draw=orange!70!red},
  ifcmodel/.style={modelbox, fill=gray!12,      draw=black!50},
  headmodel/.style={modelbox, fill=cvprblue!28, draw=cvprblue!75!black, line width=1.0pt},
  capframe/.style={draw, dashed, rounded corners=2pt, fill=gray!4, inner sep=1.6mm},
  capframeV/.style={draw, dashed, rounded corners=2pt, fill=cvprblue!8, inner sep=1.6mm},
  capout/.style={draw, line width=0.7pt, rounded corners=3pt,
                 fill=orange!22, draw=orange!70!red, align=left, font=\tiny\bfseries,
                 inner sep=2mm,
                 drop shadow={shadow xshift=0.3mm, shadow yshift=-0.3mm, opacity=0.2}},
  bigarr/.style={->, line width=0.9pt, draw=black!70, >=Latex},
  thinarr/.style={->, line width=0.7pt, dashed, draw=gray!70, >=Latex},
  stagetag/.style={fill=cvprblue!30!black, text=white, font=\tiny\bfseries,
                   rounded corners=2pt, inner sep=2pt},
  stagetagT/.style={fill=red!70!orange, text=white, font=\tiny\bfseries,
                    rounded corners=2pt, inner sep=2pt},
  capcard/.style={draw, line width=0.4pt, fill=white, rounded corners=0.5pt, font=\tiny,
                  inner sep=1.2pt, anchor=west, text width=21mm},
]

\node[imgbox] (img) at (0,0) {\textit{image}\\$I$};
\node[encmodel,  right=12mm of img]  (bigG) {\\\textbf{Retrieval Encoder}\,\frozenicon\\\scriptsize\itshape \mdseries frozen dual encoder};
\node[blipmodel, right=14mm of bigG] (blip) {\\\textbf{Verifier}\,\frozenicon\\\scriptsize\itshape \mdseries frozen cross-attention matcher};
\node[ifcmodel,  right=14mm of blip] (ifc)  {\\\textbf{Captioner}\,\frozenicon\\\scriptsize\itshape \mdseries +\,prompt $\mathcal{R}^\star_5$};
\node[headmodel, right=14mm of ifc]  (csfn) {\\\textbf{TriFuse $+$ MemAttend}\,\trainedicon\\\scriptsize\itshape \mdseries +\,memory $\mathcal{R}^\star_5$};

\node[stagetag,  anchor=north west] at ($(bigG.north west)+(1.5mm,-1.5mm)$) {Stage 1};
\node[stagetag,  anchor=north west] at ($(blip.north west)+(1.5mm,-1.5mm)$) {Stage 2};
\node[stagetag,  anchor=north west] at ($(ifc.north west) +(1.5mm,-1.5mm)$) {Stage 3};
\node[stagetagT, anchor=north west] at ($(csfn.north west)+(1.5mm,-1.5mm)$) {Stage 4};

\draw[bigarr] (img.east)  -- (bigG.west);
\draw[bigarr] (bigG.east) -- (blip.west);
\draw[bigarr] (blip.east) -- (ifc.west);
\draw[bigarr] (ifc.east)  -- (csfn.west);

\node[capout, right=14mm of csfn, text width=44mm] (CAP) {%
  $\widehat{y}(I)$:\\
  ``A man riding a motorcycle on a dirt road on a hill side.''};
\draw[bigarr] (csfn.east) -- (CAP.west);

\begin{scope}[shift={($(bigG.north)+(0,21mm)$)}]
  \fill[yellow!22, draw=yellow!50!brown, line width=0.5pt,
        drop shadow={shadow xshift=0.3mm, shadow yshift=-0.3mm, opacity=0.18}]
    (-8mm,0mm)
    -- (-8mm,-6mm)
    arc[start angle=180, end angle=360, x radius=8mm, y radius=2mm]
    -- (8mm,0mm)
    arc[start angle=0,   end angle=180, x radius=8mm, y radius=2mm]
    -- cycle;
  \fill[yellow!35, draw=yellow!50!brown, line width=0.5pt]
    (0mm,0mm) ellipse [x radius=8mm, y radius=2mm];
  \draw[gray!55, line width=0.3pt]
    (-6.2mm,-2mm) arc[start angle=180, end angle=360, x radius=6.2mm, y radius=1.6mm];
  \draw[gray!55, line width=0.3pt]
    (-6.2mm,-4mm) arc[start angle=180, end angle=360, x radius=6.2mm, y radius=1.6mm];

  \node[anchor=south, font=\scriptsize\bfseries, color=black!85]
    at (0mm,2.5mm) {Memory bank};
  \node[anchor=north, font=\tiny\itshape, color=black!70, align=center]
    at (0mm,-7mm) {$\mathcal{C}_\text{train}\!=\!566$K\\caption corpus};
\end{scope}

\begin{scope}[shift={($(bigG.north)+(20mm,21mm)$)}]
  \node[capcard, rotate=-3] (c1) at (0,1mm)  {``A man on a motorcycle\,\dots''};
  \node[capcard, rotate=2,  anchor=west] (c2) at (-1.5mm,-2mm)  {``Dog catching a frisbee\,\dots''};
  \node[capcard, rotate=-1, anchor=west] (c3) at (-2.5mm,-5mm) {``A bowl of fruit\,\dots''};
  \node[font=\tiny\itshape, color=black!60, anchor=north]
    at (4mm,-7.5mm) {\dots\,$566$K captions};
\end{scope}

\draw[thinarr] ($(bigG.north)+(0,12mm)$) -- (bigG.north);
\node[font=\tiny\itshape, color=black!60, anchor=west]
  at ($(bigG.north)+(1mm,5mm)$) {index lookup};

\node[capframe, anchor=north, below=8mm of bigG, text width=42mm] {%
  \tiny\textbf{top-9 $\mathcal{R}_9$:}\\
  $\bullet$\,``A man riding a motorbike\,\dots''\\
  $\bullet$\,``A motorcycle on a hillside\,\dots''\\
  $\bullet$\,\emph{(7 more)}};
\node[capframeV, anchor=north, below=8mm of blip, text width=42mm] {%
  \tiny\textbf{top-5 $\mathcal{R}^\star_5$ (Verifier softmax):}\\
  $\bullet$\,$0.94$\;``A man on a motorcycle\,\dots''\\
  $\bullet$\,$0.91$\;``Motorcycle, dirt road\,\dots''\\
  $\bullet$\,\emph{(3 more)}};
\node[capframe, anchor=north, below=8mm of ifc, text width=42mm] {%
  \tiny\textbf{beam-20 $\mathcal{Y}_{20}$ (LM-ranked):}\\
  $y_1$:\,``A man on a motorcycle on a road''\\
  $y_2$:\,``A man riding a motorbike\,\dots''\\
  $\vdots$\\
  $y_{20}$:\,\emph{(17 more)}};
\node[capframeV, anchor=north, below=8mm of csfn, text width=42mm] {%
  \tiny\textbf{ranking score $\hat{r}_y$:}\\
  $\beta\,z(\mathrm{TriFuse})\!+\!(1{-}\beta)\,z(\mathrm{MemAttend})$\\
  $\beta=0.75$, take $\arg\max_y$};

\node[draw, line width=0.7pt, dashed, rounded corners=3pt, fill=white,
      font=\scriptsize, align=left, inner sep=3pt, anchor=north east]
  at ($(CAP.south east)+(0,-3mm)$)
  {\frozenicon\;\,\textbf{frozen} pretrained model\\
   \trainedicon\;\,\textbf{trained} in this paper (self-supervised)};

\end{tikzpicture}%
}

\begin{figure*}[t]
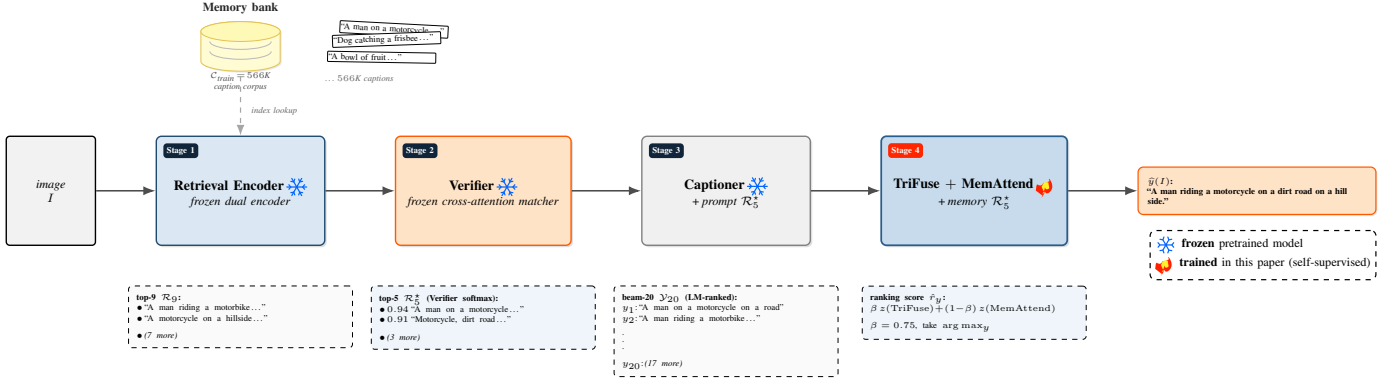

\centering
\resizebox{\textwidth}{!}{\usebox{\pipeboxA}}
\caption{\textbf{Full inference pipeline.}
The four stages run left-to-right; each agent box carries its role and any side-channel input it consumes, and the concrete output of each stage is shown in the dashed card below. The yellow \emph{Memory bank} cylinder above Stage 1 represents the external corpus of $\mathcal{C}_\text{train}\!=\!566$K training captions, with sample caption cards shown alongside, from which the Retrieval Encoder draws. \textbf{Stage 1, Retrieval}: a frozen Retrieval Encoder encodes $I$; image--text cosine over the precomputed corpus index returns the top-$9$ candidates $\mathcal{R}_9$. \textbf{Stage 2, Verification}: a frozen Cross-Attention Verifier re-ranks the $9$ candidates by image--text matching softmax and keeps the top-$5$, the Verifier-adjudicated retrieval pool $\mathcal{R}^\star_5$. \textbf{Stage 3, Captioning}: the unmodified Captioner consumes the image feature and $\mathcal{R}^\star_5$ as soft prefix and hard prompt at $K{=}3$, emitting a $20$-wide beam $\mathcal{Y}_{20}$. \textbf{Stage 4, Reranking}, the only trained stage: TriFuse (MLP) and MemAttend (transformer with $\mathcal{R}^\star_5$ as memory tokens) jointly score the beam; their $z$-normalised convex combination at $\beta{=}0.75$ selects the final caption $\widehat{y}$. TriFuse and MemAttend are trained self-supervised by Borda-consensus distillation across the three frozen scorers, namely the LM log-probability, the Retrieval cosine, and the Verifier matching score, with \textbf{no paired image--caption labels}. Only the upstream pretrained scorers, loaded frozen as indicated by the \frozenicon\ snowflake, ever saw image--text pairs.}
\label{fig:pipeline}
\end{figure*}

Existing captioner-text-only ZIC pipelines face two architectural weaknesses. \emph{First}, image--text alignment is scored at exactly one place, the initial CLIP-cosine retrieval, so the retrieval pool is dual-encoder-noisy. \emph{Second}, the Captioner's beam is selected by language-model probability alone and is therefore biased toward fluent-but-not-fully-grounded outputs. We address both by scoring image--text alignment at \emph{multiple} alignment checkpoints. Our framework integrates three components on top of an unchanged IFCap Captioner. At Stage~1 we install a stronger frozen \textbf{Retrieval Encoder}. At Stage~2 we insert a frozen \textbf{Cross-Attention Verifier} between retrieval and decoding. At Stage~4 we attach a learned \textbf{Beam Reranker} that fuses three frozen-scorer signals into a single ranking score. The section proceeds in four subsections. \Cref{sec:method_ifcap} fixes notation and reviews the IFCap baseline. \Cref{sec:method_pipeline} introduces the triadic alignment pipeline as four equations. \Cref{sec:method_csfn} introduces TriFuse, the MLP Reranker. \Cref{sec:method_macst} introduces MemAttend, the memory-augmented transformer Reranker. TriFuse and MemAttend are the only learned modules in the framework. They are trained under the protocols defined in \Cref{sec:setup_protocols}: the headline inductive protocol fits the heads on the disjoint COCO Karpathy validation beam dump and applies them frozen to test; the transductive variant fits on test-image beam features using only frozen-scorer pseudo-labels and is reported as an upper bound. The overall flow is shown in Figure~\ref{fig:pipeline}.

\subsection{Notation and the IFCap baseline}
\label{sec:method_ifcap}

Let $\mathcal{C}_\text{train} = \{c_1, \ldots, c_N\}$ be a text-only training corpus of $N \approx 566{,}747$ captions, drawn from the COCO Karpathy training split, which contains about $113$K images with 5 reference captions each. Let $\text{CLIP}_\text{img}(\cdot)$ and $\text{CLIP}_\text{txt}(\cdot)$ denote OpenAI's CLIP ViT-B/32 image and text encoders, and let $\mathbf{t}_c = \text{CLIP}_\text{txt}(c)/\|\cdot\|$ denote the L2-normalised CLIP-text feature of caption $c$. For a test image $I$, let $\hat{\boldsymbol{\imath}} = \text{CLIP}_\text{img}(I)/\|\cdot\|$.

IFCap~\cite{lee2024ifcap} trains a small mapping network $M_\theta$ on the text corpus alone, using CLIP-text features as a proxy for the image features it will see at inference. Concretely, IFCap perturbs each $\mathbf{t}_c$ with isotropic Gaussian noise of variance $\sigma^2 = 0.016$ to mimic the distribution of $\hat{\boldsymbol{\imath}}$, and trains $M_\theta(\tilde{\mathbf{t}}_c, \{\mathbf{t}_{r_1}, \ldots, \mathbf{t}_{r_k}\})$ to predict the caption $c$ given $k{=}5$ nearest-text retrievals. At inference, the noise-injected proxy $\tilde{\mathbf{t}}_c$ is replaced by the actual image feature $\hat{\boldsymbol{\imath}}$. A retrieval pool $\mathcal{R}_L(I)$ of size $L{=}9$ is built by ranking $\mathcal{C}_\text{train}$ by cosine to $\hat{\boldsymbol{\imath}}$, and the Captioner consumes $\mathcal{R}_5(I) \subset \mathcal{R}_9(I)$ in two ways: as a soft prefix~\cite{li2021prefix} through $M_\theta$, and as a hard prompt through a frequency-based entity filter that admits a noun if it appears in $\geq K{=}5$ of the $9$ retrieved captions. On COCO Karpathy test, this pipeline reaches CIDEr $108.0$.

\subsection{Triadic alignment pipeline}
\label{sec:method_pipeline}

We replace the single-checkpoint alignment scoring of IFCap with a four-stage pipeline: Stage~1 is a stronger dual-encoder retrieval, Stage~2 is a Cross-Attention Verifier, Stage~3 is the unchanged IFCap Captioner, and Stage~4 is a learned Beam Reranker. We describe each stage in turn, immediately following the equation that defines it.

\smallskip
\noindent\textbf{Stage 1: Retrieval.} For test image $I$, we retrieve the nine closest training captions in the cosine space of a strong frozen dual encoder:
\begin{equation}
\mathcal{R}_9(I) \;=\; \mathop{\text{top-9}}_{c \in \mathcal{C}_\text{train}}~~\phi_\text{img}(I)\cdot \phi_\text{txt}(c).
\label{eq:stage1}
\end{equation}
IFCap retrieves with OpenAI CLIP ViT-B/32~\cite{radford2021clip}, a $151$M-parameter dual encoder trained on $400$M web image--text pairs. In this work we instantiate the Retrieval Encoder with ViT-bigG/14~\cite{cherti2022openclip}, an approximately $2.5$B-parameter encoder trained on LAION-$2$B~\cite{schuhmann2022laion}; the image encoder, the text encoder, and the corpus index all come from this model. The dual-encoder retrieval index has not been revisited as foundation-CLIP scaling laws improved, and our ablation in Section~\ref{sec:abl_factorial} shows this single substitution is the largest single-component lift in the table, contributing $+3.6$ CIDEr. For the remainder of the paper we refer to the Stage-1 model as \emph{the Retrieval Encoder} when context is clear.

\smallskip
\noindent\textbf{Stage 2: Cross-Attention Verification.} We re-score the $9$ candidates with a frozen cross-attention image--text matcher, instantiated here with the BLIP-ITM head~\cite{li2022blip}, and keep the top $5$:
\begin{equation}
\mathcal{R}^\star_5(I) \;=\; \mathop{\text{top-5}}_{c \in \mathcal{R}_9(I)}~~s_V(I, c),
\label{eq:stage2}
\end{equation}
where $s_V(I, c) \in (0, 1)$ is the Verifier's frozen softmax matching score. Even with a stronger dual encoder at Stage 1, ranks remain imperfect: dual encoders score independent embeddings without any cross-attention. We measure the rank correlation between the Retrieval Encoder cosine and the Verifier score on the top-$9$ across all $5{,}000$ COCO Karpathy test images and obtain Spearman $\rho \!\approx\! 0.4$. The two scorers disagree on which retrieval best matches the image roughly $60\%$ of the time, and the Verifier ranking is the one downstream captioning cares about, since the cross-attention matcher was trained discriminatively for image--text alignment. We refer to the Stage-2 model as \emph{the Verifier} for the remainder of the paper. \emph{Threshold rescaling.} IFCap's entity filter admits a noun if it appears in $\geq K$ of the retrieved captions, with default $K{=}5$ over $L{=}9$ captions, an admission fraction of $5/9 \approx 56\%$. After the Verifier rerank shrinks the pool from $9$ to $5$, the same $K{=}5$ would require unanimous agreement and collapse the prompt to near-empty; we rescale to $K{=}3$ over $L{=}5$, an admission fraction of $3/5 = 60\%$. Of the available integer thresholds at $L{=}5$, $K{=}3$ is the closest to the original fraction, since $K{=}2$ would admit $40\%$. The combination of Verifier rerank \emph{and} threshold rescale is mutually enabling: neither works in isolation, as shown in rows 3--5 of \Cref{tab:ablation}.

\smallskip
\noindent\textbf{Stage 3: IFCap Captioning (unchanged).} Run the unmodified IFCap Captioner on the verified pool to emit a wide beam:
\begin{equation}
\mathcal{Y}_{20}(I) \;=\; \text{IFCap-beam}_{20}\big(\hat{\boldsymbol{\imath}},\; \mathcal{R}^\star_5(I)\big).
\label{eq:caption}
\end{equation}
Here $\hat{\boldsymbol{\imath}}$ is the ViT-B/32 image feature that IFCap's released mapping network expects, and $\text{IFCap-beam}_{20}$ runs IFCap with the rescaled entity-filter $K{=}3$ and beam width $20$. \emph{The Captioner is unchanged; only its inputs are.}

\smallskip
\noindent\textbf{Stage 4: Learned Beam Rerank.} Pick the final caption by maximising the learned Beam Reranker $\mathrm{TriFuse}_\phi$, or its convex combination with MemAttend described in \Cref{sec:method_macst}, over the $20$ candidates:
\begin{equation}
\widehat{y}(I) \;=\; \mathop{\arg\max}_{y \in \mathcal{Y}_{20}(I)}~ \mathrm{TriFuse}_\phi\!\big(\mathbf{f}_y(I)\big).
\label{eq:rerank}
\end{equation}
Each candidate is summarised by a $3$-d feature vector that fuses the three frozen scorers, the language model's log-probability, the Retrieval Encoder's cosine, and the Verifier's matching score, into a per-candidate fingerprint:
\begin{equation}
\mathbf{f}_y(I) \;=\; \Big[\,\underbrace{z(\ell_y)}_{\text{LM logprob}},\ \underbrace{z(s_R^{\,y})}_{\text{Retrieval cosine}},\ \underbrace{z(s_V(I,y))}_{\text{Verifier}}\,\Big] \in \mathbb{R}^3,
\end{equation}
with $\ell_y \!=\! \log p_\text{GPT-2}(y \mid \mathcal{R}^\star_5(I))$, $s_R^{\,y} \!=\! \cos(\phi_\text{img}(I),\, \phi_\text{txt}(y))$, and $z(\cdot)$ per-image z-normalisation across the $20$-wide beam, so the three signals share a common scale. The IFCap Captioner's beam is otherwise selected by language-model probability alone, which ranks fluency rather than visual matching; we empirically observe that a frozen image--text scorer applied at the output picks a different caption roughly $35\%$ of the time, and CIDEr improves whenever it does. $\mathrm{TriFuse}_\phi$ is the only learned module in the pipeline; we describe it next.

\subsection{TriFuse: minimal MLP head}
\label{sec:method_csfn}

\newsavebox{\csfnbox}
\savebox{\csfnbox}{%
\begin{tikzpicture}[
  font=\small,
  >={Latex[length=2.0mm,width=1.6mm]},
  imgbox/.style={draw, thick, rounded corners=1.5pt, minimum width=20mm, minimum height=9mm,
                 fill=black!6, align=center, font=\scriptsize},
  promptbox/.style={draw, line width=0.7pt, rounded corners=2pt,
                    fill=cvprblue!8, align=left, font=\scriptsize, text width=40mm,
                    inner sep=2mm,
                    drop shadow={shadow xshift=0.3mm, shadow yshift=-0.3mm, opacity=0.18}},
  ifcsub/.style={draw, line width=0.6pt, rounded corners=2pt, fill=gray!18,
                 minimum width=30mm, minimum height=7mm, align=center,
                 font=\scriptsize\bfseries},
  frozen/.style={draw, line width=0.7pt, rounded corners=2pt, minimum width=20mm, minimum height=9mm,
                 fill=gray!12, align=center, font=\scriptsize\bfseries,
                 drop shadow={shadow xshift=0.3mm, shadow yshift=-0.3mm, opacity=0.18}},
  feat/.style={draw, line width=0.6pt, circle, minimum size=7mm, fill=cvprblue!10,
               font=\scriptsize, inner sep=0pt},
  preproc/.style={draw, line width=0.6pt, rounded corners=2pt, minimum width=70mm, minimum height=7mm,
                  fill=yellow!18, align=center, font=\scriptsize},
  trainable/.style={draw, line width=1pt, rounded corners=2pt, minimum width=42mm, minimum height=7mm,
                    fill=cvprblue!22, draw=cvprblue!60!black, align=center, font=\scriptsize\bfseries,
                    drop shadow={shadow xshift=0.3mm, shadow yshift=-0.3mm, opacity=0.2}},
  argmaxbox/.style={draw, line width=0.7pt, rounded corners=2pt, minimum width=72mm, minimum height=8mm,
                    fill=orange!22, draw=orange!70!red, align=center, font=\scriptsize\bfseries,
                    drop shadow={shadow xshift=0.3mm, shadow yshift=-0.3mm, opacity=0.2}},
  trainbox/.style={draw, line width=0.7pt, dashed, rounded corners=3pt, fill=cvprblue!4, inner sep=2mm},
  arr/.style={->, line width=0.7pt, draw=black!70, >=Latex},
  thinarr/.style={->, line width=0.6pt, draw=gray!75, >=Latex},
  trainarr/.style={->, line width=0.7pt, draw=red!70!orange, dashed, >=Latex},
  arrlbl/.style={font=\tiny\itshape, color=black!70, fill=white, inner sep=1pt},
]

\node[imgbox] (img) at (0,0) {\textit{image}\,$I$};
\node[promptbox, anchor=west] (prompt) at ($(img.east)+(20mm,0)$) {%
  \tiny\textbf{prompt:} top-$5$ Verifier-adjudicated retrievals $\mathcal{R}^\star_5(I)$\\
  $\bullet$\,``A man on a motorcycle\,\dots''\\
  $\bullet$\,``Motorcycle on a hill\,\dots'' \emph{(3 more)}};

\node[ifcsub, fill=cvprblue!12, below=14mm of img]      (sub1) {ViT-B/32};
\node[ifcsub, fill=gray!22,    below=3mm of sub1]       (sub2) {mapping net $M_\theta$};
\node[ifcsub, fill=orange!22,  below=3mm of sub2]       (sub3) {entity filter ($K{=}3$)};
\node[ifcsub, fill=gray!22,    below=3mm of sub3]       (sub4) {GPT-2 decoder};

\draw[->, line width=0.6pt] (sub1.south) -- (sub2.north);
\draw[->, line width=0.6pt] (sub2.south) -- (sub3.north);
\draw[->, line width=0.6pt] (sub3.south) -- (sub4.north);

\begin{scope}[on background layer]
  \node[draw, line width=1pt, rounded corners=3pt, fill=gray!6,
        fit=(sub1)(sub4), inner xsep=4mm, inner ysep=7mm,
        drop shadow={shadow xshift=0.3mm, shadow yshift=-0.3mm, opacity=0.2}]
        (ifccont) {};
\end{scope}
\node[anchor=south, font=\footnotesize\bfseries, rotate=90]
  at ($(ifccont.west)+(-1.5mm,0)$) (ifctitle) {Captioner\,\frozenicon};

\draw[arr] (img.south) -- (sub1.north);
\coordinate (busTop) at ($(sub2.east)+(8mm,0)$);
\coordinate (busBot) at ($(sub3.east)+(8mm,0)$);
\draw[line width=0.7pt, draw=black!70]
  (prompt.south) -- ++(0,-3mm) -| (busTop);
\draw[line width=0.7pt, draw=black!70] (busTop) -- (busBot);
\draw[arr] (busTop) -- (sub2.east);
\draw[arr] (busBot) -- (sub3.east);

\node[draw, line width=0.7pt, rounded corners=2pt, minimum width=72mm, minimum height=9mm,
      fill=gray!4, align=center, font=\scriptsize, below=10mm of sub4] (beam)
  {\textbf{beam-$20$ candidates}\; $\mathcal{Y}_{20}(I)= \{y_1, y_2, \ldots, y_{20}\}$};
\draw[arr] (sub4.south) -- (beam.north -| sub4.south)
  node[arrlbl, midway, fill=white]{GPT-2 output};

\node[frozen, below=8mm of beam, xshift=-26mm] (gpt2) {GPT-2 LM\,\frozenicon};
\node[frozen, below=8mm of beam, xshift=0mm]   (bigG) {Retrieval Encoder\,\frozenicon};
\node[frozen, below=8mm of beam, xshift=26mm]  (blip) {Verifier\,\frozenicon};

\draw[arr] (beam.south) -- (gpt2.north)
  node[arrlbl, near end, fill=white]{$y$};
\draw[arr] (beam.south) -- (bigG.north);
\draw[arr] (beam.south) -- (blip.north);

\node[feat, below=3mm of gpt2] (lm) {$\ell_y$};
\node[feat, below=3mm of bigG] (sG) {$s_R^{\,y}$};
\node[feat, below=3mm of blip] (sB) {$s^{\textsc{b}}_y$};

\draw[arr] (gpt2.south) -- (lm.north);
\draw[arr] (bigG.south) -- (sG.north);
\draw[arr] (blip.south) -- (sB.north);

\node[preproc, below=4mm of sG, anchor=north] (znorm) {$z$-normalise per image (across $20$-wide beam)};
\draw[arr] (lm.south) -- (lm |- znorm.north);
\draw[arr] (sG.south) -- (sG |- znorm.north);
\draw[arr] (sB.south) -- (sB |- znorm.north);

\node[draw, line width=0.6pt, rounded corners=2pt, fill=gray!6, font=\scriptsize,
      align=center, below=3mm of znorm, minimum width=70mm, minimum height=8mm] (fy)
      {\textbf{feature vector $\mathbf{f}_y(I)\!\in\!\mathbb{R}^3$}\\\tiny\mdseries\itshape concatenated $z$-scores of the three scorers};
\draw[arr] (znorm.south) -- (fy.north);

\begin{scope}[on background layer]
  \node[draw=black!25, line width=0.4pt, rounded corners=3pt, fill=gray!2,
        fit=(img)(prompt)(ifccont)(beam)(gpt2)(blip)(fy),
        inner xsep=3mm, inner ysep=3mm] (leftpanel) {};
\end{scope}
\node[anchor=north west, font=\scriptsize\itshape, color=black!55]
  at ($(leftpanel.north west)+(2mm,-1.5mm)$) {Inference pathway (frozen)};

\coordinate (rightOrigin) at ($(prompt.east)+(38mm,-3mm)$);

\node[font=\footnotesize\bfseries, anchor=north, color=cvprblue!55!black]
  at (rightOrigin) (cstitle) {TriFuse Reranker\,\trainedicon};

\node[trainable, below=3mm of cstitle, align=center] (lin1)
  {Linear\,\trainedicon\\\tiny\mdseries\itshape lift $3\!\to\!8$};
\node[trainable, below=1.6mm of lin1, fill=cvprblue!10, align=center] (act)
  {GELU\\\tiny\mdseries\itshape non-linearity};
\node[trainable, below=1.6mm of act, align=center] (lin2)
  {Linear\,\trainedicon\\\tiny\mdseries\itshape hidden $8\!\to\!8$};
\node[trainable, below=1.6mm of lin2, fill=cvprblue!10, align=center] (act2)
  {GELU\\\tiny\mdseries\itshape non-linearity};
\node[trainable, below=1.6mm of act2, align=center] (lin3)
  {Linear\,\trainedicon\\\tiny\mdseries\itshape project $8\!\to\!1$};
\draw[arr] (lin1.south) -- (act.north);
\draw[arr] (act.south) -- (lin2.north);
\draw[arr] (lin2.south) -- (act2.north);
\draw[arr] (act2.south) -- (lin3.north);

\node[feat, below=2.5mm of lin3, fill=cvprblue!28, minimum size=9mm] (out) {$\hat{r}_y$};
\draw[arr] (lin3.south) -- (out.north);

\node[argmaxbox, below=4mm of out, minimum width=62mm, align=center]
  (argmax) {$\arg\max$ over the $20$-wide beam\\\tiny\mdseries\itshape selects final caption $\widehat{y}(I)$};
\draw[arr] (out.south) -- (argmax.north);

\draw[arr] (fy.east) -| ($(fy.east)!0.5!(lin1.west)+(0,0)$) |- (lin1.west)
  node[arrlbl, pos=0.25, fill=white, anchor=south]{$\mathbf{f}_y(I)$};

\node[anchor=north, font=\scriptsize\bfseries, color=cvprblue!55!black,
      below=4mm of argmax]
   (trainhead) {Self-supervised training (no paired labels)};

\node[draw, line width=0.6pt, rounded corners=2pt, fill=white, font=\scriptsize, align=center,
      below=3mm of trainhead, minimum width=72mm] (rank)
   {\textbf{Per-image ranks}\\\tiny\mdseries\itshape rank of $y$ under each frozen scorer ($0$--$19$)};

\node[draw, line width=0.6pt, rounded corners=2pt, fill=yellow!22, font=\scriptsize, align=center,
      below=3mm of rank, minimum width=72mm] (qy)
   {\textbf{Borda consensus}\\\tiny\mdseries\itshape average inverted ranks $\to$ softmax pseudo-label};

\node[draw, line width=0.6pt, rounded corners=2pt, fill=red!10, font=\scriptsize, align=center,
      below=3mm of qy, minimum width=72mm] (loss)
   {\textbf{Listwise cross-entropy loss}\\\tiny\mdseries\itshape trains $\phi$ to match the consensus pseudo-label};

\draw[arr] (rank.south) -- (qy.north);
\draw[arr] (qy.south)   -- (loss.north);

\begin{scope}[on background layer]
  \node[trainbox, fit=(trainhead)(rank)(qy)(loss), inner sep=3mm] (trainbg) {};
\end{scope}

\draw[trainarr] (loss.east) -- ++(6mm,0) |- (lin1.east);
\node[font=\tiny\itshape\bfseries, fill=white, draw=red!60!orange, line width=0.3pt,
      rounded corners=1pt, inner sep=1.5pt, text=red!50!black,
      anchor=west]
  at ($(loss.east)+(7mm,8mm)$) {$\nabla\mathcal{L}$ updates $\phi$};

\begin{scope}[on background layer]
  \node[draw=cvprblue!35, line width=0.4pt, rounded corners=3pt, fill=cvprblue!2,
        fit=(cstitle)(lin1)(out)(argmax)(trainbg),
        inner xsep=4mm, inner ysep=3mm] (rightpanel) {};
\end{scope}

\end{tikzpicture}%
}

\begin{figure*}[!t]
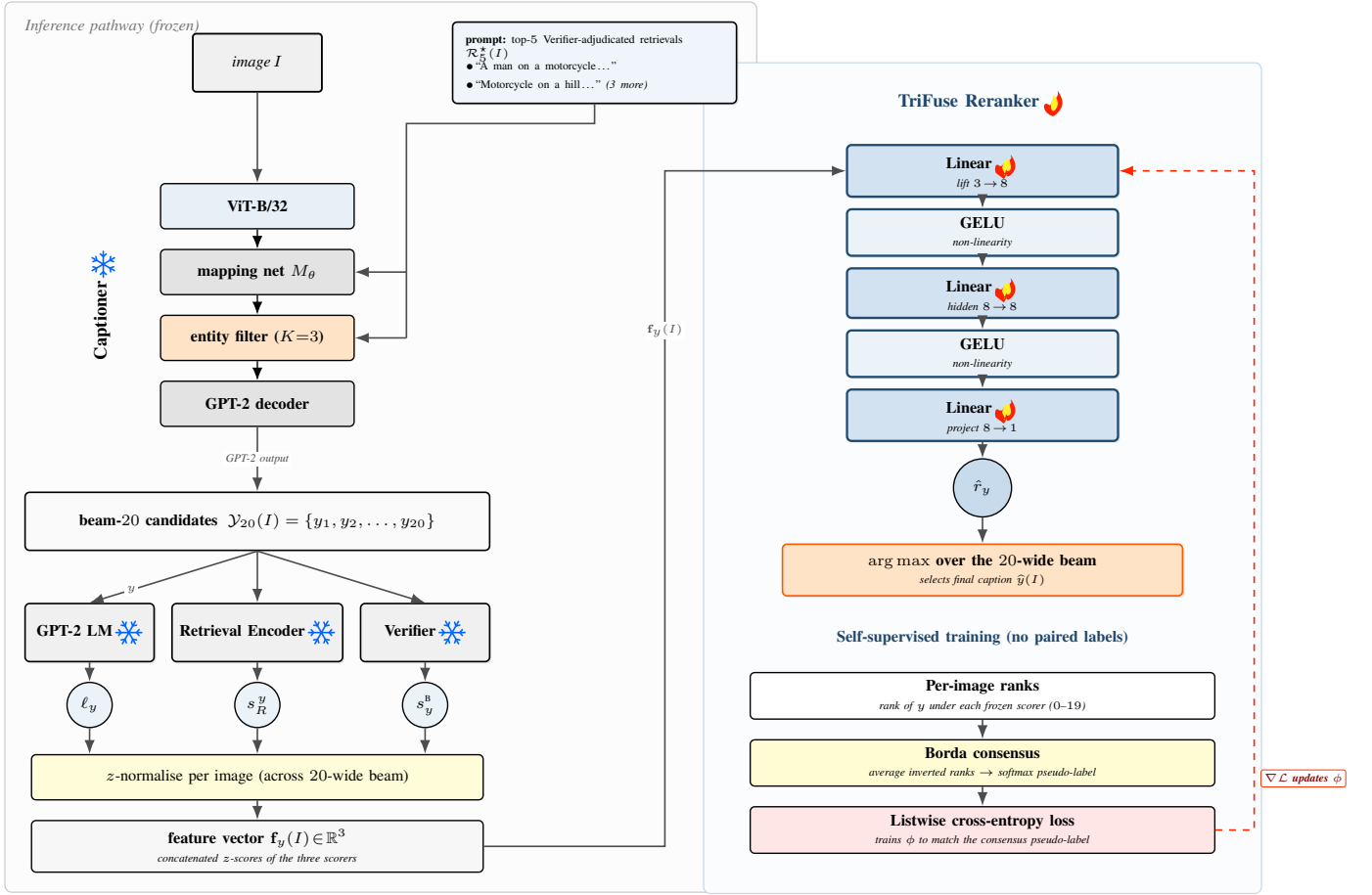

\centering
\resizebox{\textwidth}{!}{\usebox{\csfnbox}}
\caption{\textbf{TriFuse architecture and training} (two-panel layout). \textit{Left panel, inference pathway (\frozenicon\ all frozen):} the unmodified Captioner is shown as a vertical stack of its four sub-components: ViT-B/32, mapping network $M_\theta$, entity filter at $K{=}3$, and GPT-2 decoder. The image $I$ feeds into ViT-B/32; the prompt $\mathcal{R}^\star_5(I)$ feeds into $M_\theta$ and the entity filter; GPT-2 emits the $20$-wide beam $\mathcal{Y}_{20}(I)$. For each candidate $y$, three frozen scorers, namely the GPT-2 LM, the ViT-bigG/14 Retrieval Encoder, and the Verifier, emit a scalar ($\ell_y$, $s_R^{\,y}$, $s^{\textsc{b}}_y$), which are $z$-normalised across the beam and concatenated into $\mathbf{f}_y(I)\!\in\!\mathbb{R}^3$. \textit{Right panel, TriFuse Reranker (\trainedicon\ blue, the only trainable module):} Linear$\to$GELU$\to$Linear$\to$GELU$\to$Linear, $113$ parameters; the argmax over the beam selects the final caption $\widehat{y}$. \textit{Bottom (dashed blue):} self-supervised training. The pseudo-label per candidate is the Borda-consensus score across the three scorers' per-image ranks, softmaxed to a target distribution, and TriFuse is trained by listwise cross-entropy. The dashed orange arrow shows where $\nabla\mathcal{L}$ updates $\phi$. \textbf{No paired image-caption labels are used.}}
\label{fig:csfn}
\end{figure*}

\paragraph{Architecture.} TriFuse is a minimal MLP with $3$ input features per beam candidate, two hidden layers of width $h{=}8$, GELU activation~\cite{hendrycks2016gelu}, and a scalar output:
\begin{align}
\mathrm{TriFuse}_\phi(\mathbf{f}) &= W_3\,\mathrm{GELU}\!\big(W_2\,\mathrm{GELU}(W_1 \mathbf{f} + \mathbf{b}_1) + \mathbf{b}_2\big) + b_3, \\
W_1 &\in \mathbb{R}^{8\times 3},\ \ W_2 \in \mathbb{R}^{8\times 8},\ \ W_3 \in \mathbb{R}^{1\times 8}.
\end{align}
Trainable parameters: $|\phi| = 113$, namely $32$ in the input layer, $72$ in the hidden layer, and $9$ in the output layer. Inference is $O(K|\phi|)$ per image with $K{=}20$, which is sub-millisecond on a single GPU.

\paragraph{Self-supervised training.} TriFuse is trained under the protocols of \Cref{sec:setup_protocols}, using only frozen-scorer pseudo-labels (no reference captions). The headline inductive protocol fits on the disjoint COCO Karpathy validation beam-$20$ dump; the transductive protocol fits on the test beam-$20$ dump and is reported as an upper bound. For each image $I$ and each candidate $y \in \mathcal{Y}_{20}(I)$, we compute the rank $r_s(y \mid I) \in \{0, \ldots, K{-}1\}$ of $y$ within the beam under each frozen scorer $s \in \mathcal{S} = \{\text{LM}, R, V\}$. The pseudo-label is the per-image softmax of the \emph{Borda-consensus score}~\cite{dwork2001rank}:
\begin{equation}
q_y(I) = \frac{1}{|\mathcal{S}|}\sum_{s \in \mathcal{S}} (K{-}1 - r_s(y \mid I)),
\end{equation}
\begin{equation}
\tilde{q}_y(I) = \frac{\exp q_y(I)}{\sum_{y'} \exp q_{y'}(I)}.
\end{equation}
The training loss is the listwise softmax cross-entropy~\cite{cao2007listnet}:
\begin{equation}
\mathcal{L}(\phi) = -\tfrac{1}{N}\sum_{I,\,y} \tilde{q}_y(I)\,\log\,\sigma_y\!\big(\mathrm{TriFuse}_\phi(\mathbf{f}_y(I))\big),
\end{equation}
where $\sigma_y(\cdot)$ is the per-image softmax over the $K$-wide beam. We train for $100$ epochs with Adam~\cite{kingma2015adam} at learning rate $10^{-2}$. \textbf{No paired image--caption supervision is used:} the pseudo-labels are derived entirely from the frozen scorers' outputs on the Captioner's own beam, so TriFuse training does not consume any data outside the strict-ZIC text-only regime. Training takes under one minute on a single GPU.

\paragraph{Why TriFuse beats a fixed-$\alpha$ linear z-mix.} A linear mix $\alpha\,z(\ell) + (1{-}\alpha)\,z(s_R)$ uses two signals weighted by a single global hyperparameter and is constrained to lie in their span. TriFuse uses three signals, adding the Cross-Attention Verifier's matching score on $(I,y)$ that the linear mix discards, and learns a non-linear combination conditioned on per-image normalised feature interactions. Empirically, TriFuse beats the best fixed-$\alpha$ linear mix on every primary metric. The Borda-consensus pseudo-label provides a stable training target even though no candidate has a ground-truth label: the $y$ that all three frozen scorers agree on tends to be a high-CIDEr caption, and learning to predict consensus generalises to the per-image argmax.

\subsection{MemAttend: memory-augmented transformer head}
\label{sec:method_macst}

\newsavebox{\macstbox}
\savebox{\macstbox}{%
\begin{tikzpicture}[
  font=\small,
  >={Latex[length=2.0mm,width=1.6mm]},
  memtok/.style={draw, line width=0.5pt, rounded corners=1.0pt, minimum width=14mm, minimum height=8mm,
                 fill=orange!22, align=center, font=\scriptsize\bfseries},
  beamtok/.style={draw, line width=0.5pt, rounded corners=1.0pt, minimum width=14mm, minimum height=8mm,
                  fill=gray!12, align=center, font=\scriptsize\bfseries},
  emb/.style={draw, line width=0.5pt, rounded corners=1.5pt, minimum width=22mm, minimum height=7mm,
              fill=yellow!18, align=center, font=\scriptsize},
  trainable/.style={draw, line width=0.8pt, rounded corners=2pt, minimum width=70mm, minimum height=10mm,
                    fill=cvprblue!22, draw=cvprblue!60!black, align=center, font=\scriptsize\bfseries,
                    drop shadow={shadow xshift=0.3mm, shadow yshift=-0.3mm, opacity=0.18}},
  tensor/.style={draw, line width=0.5pt, rounded corners=1.5pt, minimum width=70mm, minimum height=7mm,
                 fill=gray!8, align=center, font=\scriptsize},
  ensbox/.style={draw, line width=0.7pt, rounded corners=2pt, minimum width=70mm, minimum height=9mm,
                 fill=red!12, align=center, font=\scriptsize\bfseries,
                 drop shadow={shadow xshift=0.3mm, shadow yshift=-0.3mm, opacity=0.18}},
  finalcap/.style={draw, line width=0.7pt, rounded corners=2pt, minimum width=70mm, minimum height=9mm,
                   fill=orange!22, draw=orange!70!red, align=center, font=\scriptsize\bfseries,
                   drop shadow={shadow xshift=0.3mm, shadow yshift=-0.3mm, opacity=0.18}},
  arr/.style={->, line width=0.7pt, draw=black!70, >=Latex},
  thinarr/.style={->, line width=0.5pt, draw=gray!75, >=Latex},
  lblsmall/.style={font=\tiny\itshape, color=black!70},
]

\node[lblsmall] (mlbl) at (-0.4, 2.3) {Memory tokens};
\node[lblsmall, anchor=west] (blbl) at (3.5, 2.3) {Beam tokens (top-$20$ candidates)};

\node[memtok] (m1) at (0,1) {$\mathrm{MEM}_1$};
\node[memtok] (m2) at (1.5,1) {$\mathrm{MEM}_2$};
\node[memtok] (m3) at (3.0,1) {$\ldots\mathrm{MEM}_5$};

\node[beamtok] (b1) at (5.0,1) {$\mathrm{BEAM}_1$};
\node[beamtok] (b2) at (6.5,1) {$\mathrm{BEAM}_2$};
\node[beamtok] (b3) at (8.0,1) {$\ldots$};
\node[beamtok] (b4) at (9.5,1) {$\mathrm{BEAM}_{20}$};

\node[lblsmall, below=0.15 of m2, align=center, font=\tiny\itshape]
  {$\mathbf{f}^M_j\!\in\!\mathbb{R}^5$ (Verifier, len, rank)};
\node[lblsmall, below=0.15 of b3, align=center, font=\tiny\itshape]
  {$\mathbf{f}^B_y\!\in\!\mathbb{R}^5$ (LM, Retrieval, Verifier, rank, len)};

\draw [decorate, decoration={brace, amplitude=4pt}, line width=0.5pt]
  ($(m1.north west)+(0,0.18)$) -- ($(m3.north east)+(0,0.18)$) node[midway,above=2pt,font=\tiny] {$M\!=\!5$};
\draw [decorate, decoration={brace, amplitude=4pt}, line width=0.5pt]
  ($(b1.north west)+(0,0.18)$) -- ($(b4.north east)+(0,0.18)$) node[midway,above=2pt,font=\tiny] {$K\!=\!20$};

\node[trainable, below=1.9 of $(m1)!0.5!(b4)$, minimum width=115mm] (proj)
  {\trainedicon\;Linear projection $W_\text{in}$ + role embedding};

\foreach \src in {m1, m2, m3, b1, b2, b3, b4} {
  \draw[arr] (\src.south) -- (\src |- proj.north);
}

\node[tensor, below=0.4 of proj] (h0)
  {embedded tokens $\mathbf{h}^{(0)}_t \in \mathbb{R}^{32}$, $\;t\in\{1,\dots,25\}$};
\draw[arr] (proj.south) -- (h0.north);

\node[trainable, below=0.4 of h0, fill=cvprblue!28] (enc1)
  {\trainedicon\;Encoder Layer~1\,~~\scriptsize\itshape\mdseries (4-head self-attn $+$ FFN)};
\draw[arr] (h0.south) -- (enc1.north);

\node[trainable, below=0.3 of enc1, fill=cvprblue!28] (enc2)
  {\trainedicon\;Encoder Layer~2 (same)};
\draw[arr] (enc1.south) -- (enc2.north);

\node[tensor, below=0.4 of enc2] (hL)
  {hidden states $\mathbf{h}^{(L)}_t \in \mathbb{R}^{32}$ (BEAM tokens go to head)};
\draw[arr] (enc2.south) -- (hL.north);

\node[trainable, below=0.4 of hL, fill=cvprblue!18] (outh)
  {\trainedicon\;Output head: linear $\to$ score per BEAM token};
\draw[arr] (hL.south) -- (outh.north);

\node[tensor, below=0.4 of outh, fill=cvprblue!28] (macstout)
  {$\mathrm{MemAttend}_\psi(I, y)\in\mathbb{R}$ for $y\in\mathcal{Y}_{20}$};
\draw[arr] (outh.south) -- (macstout.north);

\node[ensbox, below=0.4 of macstout]
  (ens) {Per-image $z$-normalise; $\beta\,z(\mathrm{TriFuse})\!+\!(1\!-\!\beta)\,z(\mathrm{MemAttend})$, $\;\beta\!=\!0.75$};
\draw[arr] (macstout.south) -- (ens.north);

\node[trainable, right=8mm of ens, fill=cvprblue!10, minimum width=30mm, minimum height=9mm,
      align=center, font=\scriptsize\bfseries] (csfn)
  {\trainedicon\;TriFuse$_\phi(\mathbf{f}^B_y)$\\\tiny\itshape \mdseries MLP, see Fig.~\ref{fig:csfn}};
\draw[thinarr, dashed] (csfn.west) -- (ens.east);

\node[finalcap, below=0.4 of ens]
  (cap) {$\widehat{y}(I) = \arg\max_{y\in\mathcal{Y}_{20}}\,\big[\beta\, z(\mathrm{TriFuse}_\phi) + (1\!-\!\beta)\, z(\mathrm{MemAttend}_\psi)\big]$};
\draw[arr] (ens.south) -- (cap.north);

\node[draw, line width=0.5pt, rounded corners=2pt, fill=cvprblue!4,
      anchor=west, inner sep=3mm,
      at={($(enc1.east)!0.5!(enc2.east)+(1.1,0)$)}] (attnpic) {%
\begin{tikzpicture}[scale=0.42]
  \node[font=\scriptsize\bfseries, anchor=south, align=center] at (2.5,7.5)
    {Attention readout: BEAM (query) $\!\to\!$ MEM (key)};
  \foreach \j/\lbl in {0/1,1/2,2/3,3/4,4/5} {
    \node[font=\tiny, color=orange!70!black, anchor=south] at (\j+0.5,5.18) {\lbl};
  }
  \node[font=\tiny\itshape, color=orange!70!black, anchor=south] at (2.5,5.55)
    {MEM$_j$};
  \foreach \row/\va/\vb/\vc/\vd/\ve in {
    0/0.88/0.32/0.07/0.04/0.04,
    1/0.32/0.88/0.32/0.07/0.04,
    2/0.07/0.32/0.88/0.32/0.07,
    3/0.04/0.07/0.32/0.88/0.32,
    4/0.04/0.04/0.07/0.32/0.88
  } {
    \fill[cvprblue,opacity=\va] (0,\row) rectangle ++(1,1);
    \fill[cvprblue,opacity=\vb] (1,\row) rectangle ++(1,1);
    \fill[cvprblue,opacity=\vc] (2,\row) rectangle ++(1,1);
    \fill[cvprblue,opacity=\vd] (3,\row) rectangle ++(1,1);
    \fill[cvprblue,opacity=\ve] (4,\row) rectangle ++(1,1);
    \foreach \col in {0,...,4} {
      \draw[gray!25, very thin] (\col,\row) rectangle ++(1,1);
    }
  }
  \draw[orange!70!red, line width=0.9pt] (0,2) rectangle (5,3);
  \node[font=\tiny, color=gray!65!black, anchor=east] at (-0.12,0.5) {$y_1$};
  \node[font=\tiny, color=orange!70!red,  anchor=east] at (-0.12,2.5) {$y_t$};
  \node[font=\tiny, color=gray!65!black, anchor=east] at (-0.12,4.5) {$y_K$};
  \draw[decorate, decoration={brace, mirror, amplitude=2.5pt},
        color=black!35, line width=0.4pt]
    (-0.65,0) -- (-0.65,5)
    node[midway, left=2.5pt, rotate=90, anchor=south,
         font=\tiny\itshape, color=black!55] {BEAM $y_t$ (query)};
  \node[font=\tiny\itshape, color=black!55, anchor=north, align=center] at (2.5,-0.22)
    {dark $=$ high attn weight \quad
     \textcolor{orange!70!red}{orange row} $=$ one $y_t$ example};
\end{tikzpicture}%
};
\draw[thinarr, dashed]
  (attnpic.west) -- ($(enc1.east)!0.5!(enc2.east)$)
  node[midway, above=1pt, font=\tiny\itshape, color=black!50] {inside encoder};

\node[draw, line width=0.5pt, dashed, rounded corners=2pt, fill=white,
      below=0.4 of cap, anchor=north west, xshift=2mm,
      inner sep=2mm, font=\scriptsize, align=left] (legend)
   {\trainedicon\;\textbf{trainable} ($|\psi|\!=\!17.4$K params):\\
    \;\;$W_\text{in}$, $\mathbf{r}_{\bullet}$, Encoder~1$/$2, $W_\text{out}$, $b_\text{out}$\\[0.5mm]
    \frozenicon\;\textbf{frozen}: all upstream scorers + Captioner};

\end{tikzpicture}%
}

\begin{figure*}[!t]
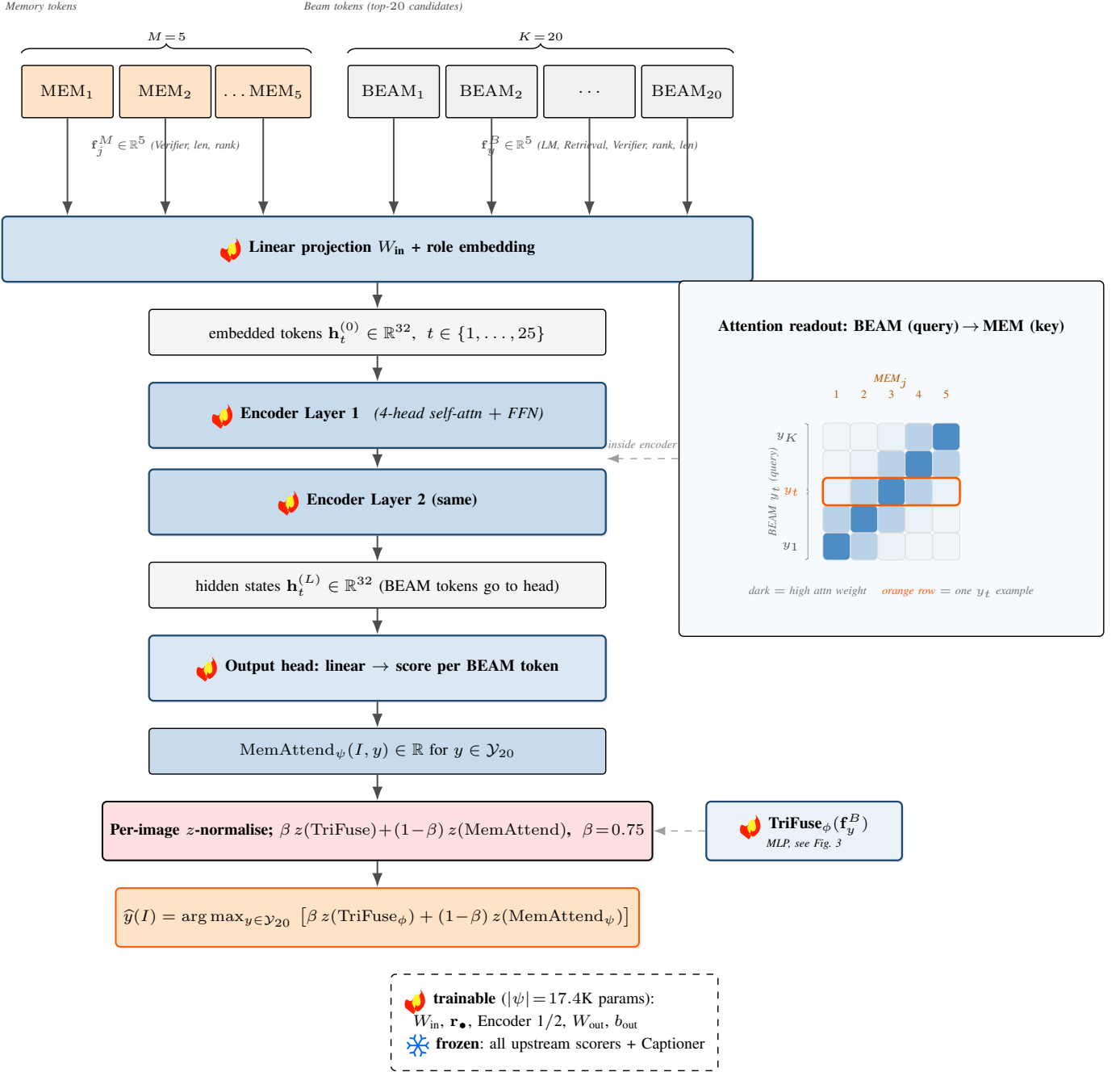

\centering
\resizebox{\textwidth}{!}{\usebox{\macstbox}}
\caption{\textbf{MemAttend architecture (Memory-Attended Reranker).}
\textit{Top:} $5$ memory tokens $\mathrm{MEM}_j$ (orange, scalars from the Stage-2 Verifier-adjudicated retrievals) and $20$ beam tokens $\mathrm{BEAM}_y$ (gray, scalars from the three frozen scorers). \textit{Stack (\trainedicon\ blue, the only trainable parameters):} a linear projection $W_\text{in}\!\in\!\mathbb{R}^{32\times 5}$ with two role embeddings (MEM vs.\ BEAM) lifts each token to $d_\text{model}{=}32$; two transformer encoder layers ($4$-head self-attention, FFN $32\!\to\!64\!\to\!32$, GELU); an output head emits a scalar per BEAM token. $|\psi|\!=\!17.4$K total. \textit{Ensemble} (red): MemAttend scores are per-image $z$-normalised and convex-combined with the parallel TriFuse head (Fig.~\ref{fig:csfn}) at $\beta{=}0.75$ to pick the final caption $\widehat{y}$. \textit{Bottom-left:} cross-attention pattern over the $K{\times}M$ token grid; each BEAM token attends to all MEM tokens (and to other BEAM tokens), implementing an explicit working-memory readout. \textit{Bottom-right:} trainability legend. Trained self-supervised by Borda-consensus distillation across the three frozen scorers; no paired image-caption labels.}
\label{fig:macst}
\end{figure*}

TriFuse's per-candidate MLP cannot directly model interactions between beam candidates or between beam candidates and the Stage-$2$ retrieval pool. We propose a complementary head with that capacity: \textbf{MemAttend}, a memory-augmented cross-scorer transformer~\cite{vaswani2017transformer} that frames the beam-rerank step as conditioning on an explicit working memory of the Verifier-adjudicated retrievals.

\paragraph{Token layout.} Each test image gives rise to a sequence of $M{+}K = 5{+}20 = 25$ tokens:
\begin{equation}
\underbrace{[\,\mathrm{MEM}_1,\ldots,\mathrm{MEM}_5\,]}_{\text{retrieval memory}}\ \Vert\ \underbrace{[\,\mathrm{BEAM}_1,\ldots,\mathrm{BEAM}_{20}\,]}_{\text{Captioner beam}}.
\end{equation}
Each beam token $\mathrm{BEAM}_y$ carries the same $5$-d feature vector TriFuse would see, namely the LM log-probability, the Retrieval Encoder's cosine, the Verifier's matching score on $(I,y)$, the beam rank, and the length. Each memory token $\mathrm{MEM}_j$ carries an analogous $5$-d feature vector for the $j$-th Verifier-verified retrieval, namely its Verifier matching score, retrieval rank, and length, with zero-padding to dimension $5$. All features are z-normalised per image. A learned role embedding $\mathbf{r} \in \mathbb{R}^{2 \times d}$ distinguishes memory tokens from beam tokens; it is added after the input projection so the model can tell the two token classes apart.

\paragraph{Architecture.} MemAttend is a $2$-layer transformer encoder with $4$ heads, hidden dimension $d_\text{model}{=}32$, feed-forward network (FFN) expansion factor $2$, GELU activation~\cite{hendrycks2016gelu}, no dropout, and a per-token scalar score head:
\begin{equation}
\mathbf{h}^{(0)} = W_\text{in}\,\mathbf{f}_t + \mathbf{r}_{\text{role}(t)}, \qquad
\mathbf{h}^{(\ell+1)} = \text{TFLayer}^{(\ell)}(\mathbf{h}^{(\ell)}),
\end{equation}
\begin{equation}
\mathrm{MemAttend}_\psi(I, y) = W_\text{out}\,\mathbf{h}^{(L)}_y + b_\text{out},
\end{equation}
where only beam tokens are scored at the output. Trainable parameters: $|\psi| = 17{,}377$, namely $192$ in the input projection, $64$ in the two role embeddings, $17{,}088$ across the two encoder layers, and $33$ in the output head. Added to TriFuse's $|\phi| = 113$, the two learned heads hold $113 + 17{,}377 = 17{,}490$ trainable parameters in total. Every other component of the pipeline is frozen, so this is the complete trainable-parameter budget of the method. Training uses the same Borda-consensus pseudo-label as TriFuse, with listwise softmax cross-entropy, Adam~\cite{kingma2015adam} at learning rate $3{\times}10^{-3}$, and $200$ epochs. \textbf{No paired image--caption supervision.}

\paragraph{Why MemAttend is not a strict superset of TriFuse.} The transformer's cross-attention lets a beam candidate's score depend on the entire memory and beam context, not just its own three frozen-scorer signals. In practice, MemAttend's higher capacity drives its training loss substantially below TriFuse's, $1.27$ versus $1.44$ over the same Borda target, but its CIDEr lands $0.2$ below TriFuse's: the transformer fits the imperfect pseudo-label more tightly and overfits, while TriFuse's tighter inductive bias generalises slightly better on its own. Because the two heads' top-$1$ picks agree on only $75$--$82\%$ of test images, they are partly orthogonal. We therefore use a fixed convex combination of their z-normalised scores,
\begin{equation}
\begin{aligned}
\widehat{y} \;=\; \arg\max_y\, \big[\,
  &\beta\, z\!\left(\mathrm{TriFuse}_\phi(\mathbf{f}_y)\right) \\
  +\;& (1-\beta)\, z\!\left(\mathrm{MemAttend}_\psi(\mathbf{f}_y, \mathcal{M})\right) \,\big],
\end{aligned}
\end{equation}
which lifts every primary metric over either head alone. MemAttend contributes most where the consensus signal is ambiguous, that is on low-margin beam candidates; TriFuse contributes most where the consensus is sharp. We discuss the per-benchmark behaviour of $\beta$ in our cross-domain experiments.

\FloatBarrier

\section{Experimental Setup}
\label{sec:setup}

\subsection{Benchmarks and metrics}
We evaluate on three benchmarks. \textbf{COCO Karpathy test}~\cite{lin2014coco,karpathy2015deepvs} contains $5{,}000$ unique images and $25{,}010$ reference captions under the IFCap convention; it serves as the primary in-domain benchmark, and the Captioner was trained on the COCO Karpathy training caption corpus alone, with no paired image--caption supervision. \textbf{Flickr30k Karpathy test}~\cite{young2014flickr30k,plummer2017flickr30k} contains $1{,}000$ images and $5{,}000$ references, and serves as a cross-domain benchmark for the COCO-trained pipeline. \textbf{NoCaps val}~\cite{agrawal2019nocaps} contains $4{,}500$ images and tests novel-object transfer, split into in-, near-, and out-domain partitions. Metrics are BLEU-$4$~\cite{papineni2002bleu}, METEOR~\cite{banerjee2005meteor}, ROUGE-L~\cite{lin2004rouge}, CIDEr~\cite{vedantam2015cider}, and SPICE~\cite{anderson2016spice} from the standard COCO Captions evaluation suite~\cite{chen2015cococaptions}; we report all metrics on a $\times 100$ scale throughout. SPICE requires Java 11; the runtime is documented in our release.

\subsection{Hardware and software}
Headline experiments run on a single NVIDIA L4 GPU with $24$ GB of memory on a GCP instance running CUDA 12 and PyTorch $2.4.1$, matching IFCap's pinned configuration. Cross-platform reproduction uses a single AMD Instinct MI355X with the gfx950 architecture, ROCm 7.2, and PyTorch $2.8.0$+rocm$6.4$. The two platforms reproduce CIDEr and SPICE to the one-decimal precision reported throughout this paper; byte-exact JSON identity holds on the L4 but not on the MI355X, because ROCm/CUDA non-determinism is amplified by the gfx942-on-gfx950 kernel override. All frozen scorer checkpoints are exact pinned versions: the OpenCLIP~\cite{cherti2022openclip} ViT-bigG/14 backbone pretrained on LAION-2B; the BLIP-ITM image--text matching head~\cite{li2022blip} fine-tuned on COCO captions, used as the Stage-2 Verifier; OpenAI CLIP ViT-B/32~\cite{radford2021clip} for the Captioner's image encoder, so the released IFCap~\cite{lee2024ifcap} mapping network is consumed unchanged; and GPT-2 small~\cite{radford2019gpt2} with 124M parameters for the Captioner's language model. The IFCap Captioner checkpoint is the released in-domain COCO model from the official IFCap repository, used unmodified. Random seed is $42$ throughout.

\subsection{Baseline reproduction}
Before measuring our pipeline we re-ran IFCap's released checkpoint, as shown in \Cref{tab:reproduction}. Every COCO and NoCaps metric reproduces the published value exactly, and the COCO in-domain generated-caption file is byte-identical to the authors' released JSON. Flickr30k CIDEr reproduces at $64.5$ against the published $64.4$, a $0.1$ gap on the $\times 100$ scale that sits at the level of the determinism drift reported in \Cref{sec:exp_robust}. Except where noted, comparisons in this paper are versus this reproduction rather than paper-quoted numbers.

\begin{table}[t]
\centering
\setlength{\tabcolsep}{4pt}
\resizebox{\columnwidth}{!}{%
\begin{tabular}{lllll}
\toprule
Benchmark & Split & Metric & Paper & Ours \\
\midrule
COCO Karp. & in-domain & CIDEr & $\mathbf{108.0}$ & $\mathbf{108.0}$ \\
COCO Karp. & in-domain & SPICE & $20.3$ & $20.3$ \\
Flickr30k Karp. & in-domain & CIDEr & $64.4$ & $64.5$ \\
NoCaps val & overall & CIDEr & $74.0$ & $74.0$ \\
NoCaps val & in/near/out & CIDEr & $70.1/72.5/72.1$ & $70.1/72.5/72.1$ \\
\bottomrule
\end{tabular}%
}
\caption{\textbf{Reproduction of IFCap from the released checkpoint.} All entries are $\times 100$, at the one-decimal precision at which IFCap publishes them. Every row matches the published value except Flickr30k CIDEr, which lands $0.1$ higher, at the run-to-run determinism level quantified in \Cref{sec:exp_robust}.}
\label{tab:reproduction}
\end{table}

\subsection{Evaluation protocols for the learned Reranker}
\label{sec:setup_protocols}
The TriFuse and MemAttend Rerankers introduce a learned component over the Captioner's beam. Because the choice of \emph{which images supply the training features} for these heads affects how the result should be interpreted, we report three protocols on the same pipeline:
\begin{enumerate}[leftmargin=1.5em,itemsep=0.15em]
\item \textbf{Training-free fixed fusion, the inductive baseline with no learned step.} The Stage-$4$ ranker is a fixed-$\alpha$ linear z-mix of LM log-prob and ViT-bigG/14 CLIPScore with $\alpha{=}0.48$, selected by Borda-consensus pseudo-label loss on a held-out beam dump and never on reference captions. No parameters are fit on test-set features; this is our strictest non-transductive baseline.
\item \textbf{Inductive learned Reranker, headline protocol.} TriFuse and MemAttend are trained on beam-$20$ features extracted from the COCO Karpathy \emph{validation} split, which contains $5{,}000$ images disjoint from the test split, using the same Borda-consensus pseudo-label, then applied unchanged to the Karpathy test beam-$20$ features. No test-set features enter training. The choice of $\alpha$, $\beta$, hidden width, and epoch count is fixed once on validation by pseudo-label loss rather than by reference CIDEr.
\item \textbf{Transductive learned Reranker, upper-bound protocol.} TriFuse and MemAttend are trained on beam-$20$ features extracted from the Karpathy \emph{test} split, with the same Borda-consensus pseudo-label, then applied to the same images. No reference captions are used as labels; the training signal is the frozen scorers' rankings on the Captioner's own beam, computed at inference time. This is a form of unsupervised test-time adaptation: parameters are fit to test inputs but no test labels are seen. We report this only as an upper bound; the inductive protocol (item 2) is our headline.
\end{enumerate}

\subsection{Hyperparameter selection criterion}
\label{sec:setup_hpsel}
The learned-component hyperparameters, namely the ensemble weight $\beta{=}0.75$, the fixed-fusion weight $\alpha{=}0.48$, the TriFuse hidden width $h{=}8$, the MemAttend hidden dimension $d_\text{model}{=}32$ with $2$ encoder layers, and the BLIP top-$k{=}5$, were selected by minimising Borda-consensus pseudo-label loss on a held-out beam dump, \emph{not} by reference-metric CIDEr. The entity-filter threshold $K{=}3$ was fixed algebraically rather than swept: the BLIP rerank shrinks the entity pool from $L{=}9$ to $L{=}5$, and $K{=}3$ admits $3/5 = 60\%$, the closest available integer threshold to IFCap's original $5/9 \approx 56\%$. We do not perform random restart sweeps; the seed is fixed at $42$.

\subsection{Computational budget}
A one-time preprocessing step encodes all $N = 566{,}747$ COCO Karpathy training captions with OpenCLIP ViT-bigG/14 and caches the L$_2$-normalised $1280$-d embeddings on disk at about $3$ GB. The encode cost is approximately $5$ GPU-hours on a single NVIDIA L4 with $24$ GB of memory, batch $128$, FP$32$; wall-clock time on faster GPUs such as the A$100$, H$100$, or MI$300$ is several times lower. The cache is reusable across all subsequent runs of the pipeline. Per test image, the additional cost over vanilla IFCap inference breaks down as follows. Stage~1, the ViT-bigG/14 image encode and top-$9$ cosine over the corpus, costs approximately $0.05$ s. Stage~2, the BLIP-ITM rerank pass over $9$ retrievals, costs around $1.0$ s. Stage~4 input features, namely one ViT-bigG/14 image encode, $20$ text encodes, and $20$ BLIP-ITM forwards, cost roughly $1.4$ s. TriFuse's own forward is sub-millisecond. The total is about $2.5$ s per image on top of vanilla IFCap inference, on a single GPU, with no Captioner retraining. TriFuse training is a one-time fit of under one minute on the $5$K-image beam-$20$ feature dump; MemAttend training takes approximately $5$ minutes over $200$ epochs.

\section{Experiments}
\label{sec:exp}

\subsection{Implementation details}
\label{sec:exp_impl}

We use the released IFCap Captioner without retraining. The only inference-time changes are the following. \emph{(i)} One-time, we encode all $566{,}747$ COCO Karpathy training captions with the frozen Stage-1 Retrieval Encoder and cache the L$_2$-normalised $1280$-d embeddings. \emph{(ii)} For each test image we encode it with the Retrieval Encoder and retrieve the top-$9$ training captions by cosine. \emph{(iii)} The frozen Verifier re-ranks the $9$ candidates and keeps the top-$5$. \emph{(iv)} The unmodified IFCap Captioner runs with entity-filter $K{=}3$ and beam width $20$ to dump a $20$-wide beam. \emph{(v)} For each beam candidate $y$ we compute three per-candidate features, the LM log-probability $\ell_y$, the Retrieval Encoder's cosine $s_R^{\,y}$, and the Verifier's matching score $s_V(I,y)$, and z-normalise per image. \emph{(vi)} TriFuse is an MLP of $113$ parameters with two hidden layers of width $h{=}8$ and GELU~\cite{hendrycks2016gelu}, trained for $100$ epochs with Adam~\cite{kingma2015adam} at learning rate $10^{-2}$ using listwise softmax cross-entropy against the Borda-consensus pseudo-label; MemAttend is a $2$-layer transformer encoder of $17.4$K parameters, trained for $200$ epochs with Adam~\cite{kingma2015adam} at learning rate $3{\times}10^{-3}$ against the same target. \emph{(vii)} We take the convex combination $\beta\,z(\mathrm{TriFuse})+(1{-}\beta)\,z(\mathrm{MemAttend})$ at $\beta{=}0.75$ and use its argmax per image as the final caption. No Captioner retraining and no weight modification of any frozen model.

\paragraph{Inference cost.} The one-time corpus encode is approximately $5$ GPU-hours on a single NVIDIA L4 (matching the budget reported in \Cref{sec:setup}). Per test image, the additional cost over vanilla IFCap is about $0.05$ s for Retrieval Encoder lookup, approximately $1.0$ s for Verifier rerank, around $1.4$ s for beam-feature extraction, and sub-millisecond for the TriFuse and MemAttend forwards, for a total of roughly $2.5$ s per image. TriFuse training is a one-time fit of under one minute; MemAttend training takes about $5$ minutes. Both heads are fit on the $5{,}000$-image beam-$20$ feature dump.

\paragraph{Hardware.} See \Cref{sec:setup} for the canonical hardware specification: headline experiments use a single NVIDIA L4 ($24$ GB, CUDA 12, PyTorch $2.4.1$, IFCap's pinned configuration); cross-platform reproduction uses a single AMD Instinct MI355X (gfx950, ROCm $7.2$, PyTorch $2.8.0$+rocm$6.4$). All reported numbers reproduce to the one-decimal CIDEr/SPICE precision used throughout on either platform.

\subsection{Datasets and metrics}
\label{sec:exp_data}

Datasets, metrics, reproduction, and protocols follow \Cref{sec:setup}. Reference-free grounding is additionally measured by ViT-bigG/14 CLIPScore, and hallucination by CHAIR~\cite{rohrbach2018chair} computed against the $5$ reference captions per test image.

\subsection{Baselines}
\label{sec:exp_baselines}

We compare against three families: (i) \emph{strict text-only ZIC}, including ZeroCap~\cite{tewel2022zerocap}, MAGIC~\cite{su2022magic}, CapDec~\cite{nukrai2022capdec}, DeCap~\cite{li2023decap}, ViECap~\cite{fei2023viecap}, MeaCap~\cite{zeng2024meacap}, and IFCap~\cite{lee2024ifcap}; (ii) \emph{strict ZIC with training-time augmentation by synthetic images}, namely SynTIC~\cite{liu2024syntic}, PCM-Net~\cite{zeng2024pcmnet}, and NES~\cite{lu2026negative}; and (iii) \emph{paired-supervised} models reported for context, namely ClipCap~\cite{mokady2021clipcap}, BLIP~\cite{li2022blip}, and BLIP-2~\cite{li2023blip2}. We also include a foundation multimodal large language model (MLLM), Qwen2.5-VL-3B~\cite{bai2025qwen25vl}, fed our retrievals as in-context exemplars. Following template practice, we keep these regimes in separate table blocks.

\subsection{Main results: COCO Karpathy}
\label{sec:exp_main}

\Cref{tab:scoreboard} reports the full metric set, with all comparators grouped by training regime. Our framework reaches \textbf{CIDEr $\mathbf{117.6}$ / SPICE $\mathbf{21.9}$}, the best strict-text-only result on this benchmark to our knowledge, surpassing the previous strict-ZIC state of the art, IFCap (CIDEr $108.0$), by $\mathbf{+9.6}$ and the strongest synthetic-image-augmented method NES (CIDEr $109.9$) by $+7.7$ without any Captioner retraining. SPICE $21.9$ further surpasses ClipCap's $21.1$, a paired-supervised baseline trained on full COCO image--caption pairs. Every reference-based metric improves simultaneously over IFCap; no metric is traded for another. The TriFuse+MemAttend ensemble contributes $+1.8$ CIDEr over a fixed-$\alpha$ z-mix that has access to the same input scorers; these are the only learned components in the pipeline.

\begin{figure}[t]
  \centering
  \includegraphics[width=\columnwidth]{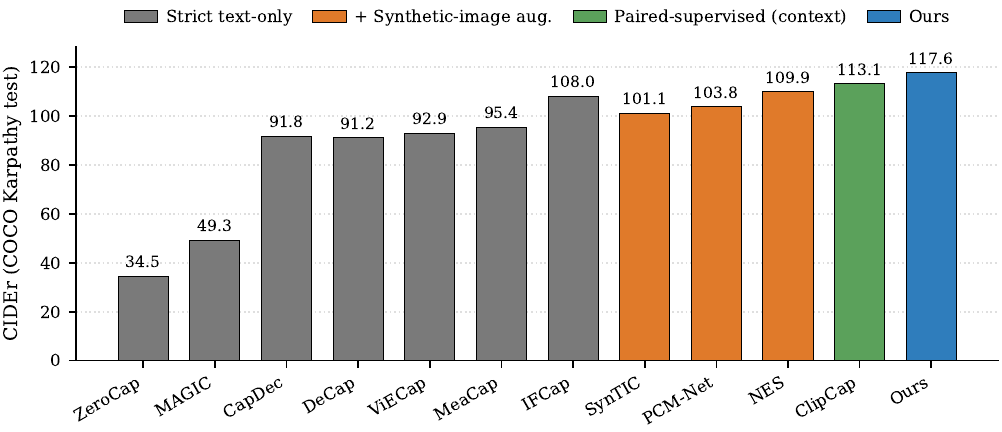}
  \caption{CIDEr on COCO Karpathy test, grouped by training regime. Progress in the captioner-text-only ZIC regime has plateaued near IFCap's $108.0$; the only entry to surpass it, NES at $109.9$, does so by stepping outside the regime via Stable-Diffusion-generated synthetic images. Our framework reaches $\mathbf{117.6}$ without retraining the Captioner and without any paired image--caption supervision, exceeding even the paired-supervised baseline ClipCap ($113.1$). Numbers from \Cref{tab:scoreboard}.}
  \label{fig:perf_coco}
\end{figure}

\begin{table*}[t]
\centering\small
\setlength{\tabcolsep}{3.5pt}
\resizebox{\textwidth}{!}{%
\begin{tabular}{lllcccccc}
\toprule
Method & Venue & Encoder & Train data & B-4 & MET & R-L & \textbf{CIDEr} & \textbf{SPICE} \\
\midrule
\multicolumn{9}{l}{\emph{Captioner-text-only ZIC; frozen paired-pretrained scorers allowed; no retraining-time image augmentation:}} \\
ZeroCap~\cite{tewel2022zerocap}    & CVPR'22         & ViT-B/32 & GPT-2 prior & ---  & 15.4 & ---  & 34.5  & 9.2  \\
MAGIC~\cite{su2022magic}           & arXiv'22        & ViT-B/32 & text only   & 12.9 & 17.4 & ---  & 49.3  & 11.3 \\
CapDec~\cite{nukrai2022capdec}     & EMNLP'22        & ViT-B/32 & text only   & 26.4 & 25.1 & ---  & 91.8  & 11.9 \\
DeCap~\cite{li2023decap}           & ICLR'23         & ViT-B/32 & text only   & 24.7 & 25.0 & ---  & 91.2  & 18.7 \\
ViECap~\cite{fei2023viecap}        & ICCV'23         & ViT-B/32 & text only   & 27.2 & 24.8 & ---  & 92.9  & 18.2 \\
MeaCap~\cite{zeng2024meacap}       & CVPR'24         & ViT-B/32 & text only   & 27.2 & 25.3 & ---  & 95.4  & 19.0 \\
IFCap~\cite{lee2024ifcap}          & EMNLP'24        & ViT-B/32 & text only   & 30.7 & 26.6 & 53.8 & 108.0 & 20.3 \\
\midrule
\multicolumn{9}{l}{\emph{Captioner-text-only ZIC with synthetic-image training augmentation:}} \\
SynTIC~\cite{liu2024syntic}        & AAAI'24         & ViT-B/32 & text + SD synth img & 29.9 & 25.8 & 53.2 & 101.1 & 19.3 \\
PCM-Net~\cite{zeng2024pcmnet}      & ECCV'24         & ViT-B/32 & text + SD synth img & 31.5 & 25.9 & 53.9 & 103.8 & 19.7 \\
NES~\cite{lu2026negative}               & AAAI'26         & ViT-B/32 & text + SD synth img & 30.8 & 26.8 & ---  & 109.9 & 20.6 \\
\midrule
\multicolumn{9}{l}{\emph{Supervised image captioning (paired image--text training; reported for context, not regime-comparable):}} \\
ClipCap~\cite{mokady2021clipcap}   & arXiv'21        & ViT-B/32 & paired       & 33.5 & 27.5 & ---  & 113.1 & 21.1 \\
BLIP~\cite{li2022blip}             & ICML'22         & ViT-B/16 & 14M paired   & 39.7 & ---  & ---  & 133.3 & ---  \\
BLIP-2~\cite{li2023blip2}          & ICML'23         & ViT-G    & 129M paired  & 43.7 & ---  & ---  & 145.8 & ---  \\
Qwen2.5-VL-3B~\cite{bai2025qwen25vl} + our retrievals & arXiv'25 & Qwen-ViT-L & MLLM paired pretrain & 21.8 & 28.1 & 50.6 & 77.1  & 23.3 \\
\midrule
\multicolumn{9}{l}{\emph{Ours: captioner-text-only ZIC under frozen paired-pretrained scorers; inference-time intervention on IFCap's released checkpoint:}} \\
\rowcolor{cvprblue!10}
\textbf{Ours (ViT-bigG/14 in + BLIP Verifier + TriFuse$+$MemAttend out, inductive)} & --- & ViT-bigG/14 & captioner text only; frozen paired scorers & $\mathbf{32.8}$ & $\mathbf{27.9}$ & $\mathbf{55.5}$ & $\mathbf{117.6}$ & $\mathbf{21.9}$ \\
\bottomrule
\end{tabular}%
}
\caption{COCO Karpathy test. CIDEr and SPICE reported $\times 100$ following prior convention. Our $117.6$ CIDEr is the best captioner-text-only ZIC result on this benchmark under frozen paired-pretrained scorers, surpassing the previous strict-ZIC state of the art, IFCap, by $+9.6$ and the strongest synthetic-image-augmented method NES by $+7.7$ without any Captioner retraining. The ``Ours'' row reports the \emph{inductive} protocol (Rerankers fit on the disjoint COCO Karpathy validation beam dump and applied frozen to test); the transductive variant differs by at most $0.1$ on any metric, and not at all on CIDEr, and is reported in \Cref{tab:protocols}; the only learned components are TriFuse ($113$ parameters, ${<}1$ min fit) and MemAttend ($17.4$K parameters, ${\approx}5$ min fit), both fit self-supervised by Borda-consensus distillation across three frozen scorers. SPICE $21.9$ surpasses ClipCap ($21.1$), a paired-supervised baseline. The IFCap row reports our reproduction of the released checkpoint (\Cref{tab:reproduction}), which is what every comparison in this paper is made against (\Cref{sec:setup}), so it also carries a ROUGE-L value that the original paper does not report. In the train-data column, SD denotes Stable Diffusion. ``Encoder'' refers to the dominant frozen image--text scorer used at retrieval and output; the Captioner itself is IFCap's released ViT-B/32 checkpoint, unmodified, in every row of this paper. Supervised methods are listed for context to bound how far the strict-ZIC regime is from the paired-data ceiling---not as direct comparators.}
\label{tab:scoreboard}
\end{table*}

\begin{figure*}[t]
  \centering
  \includegraphics[width=\textwidth]{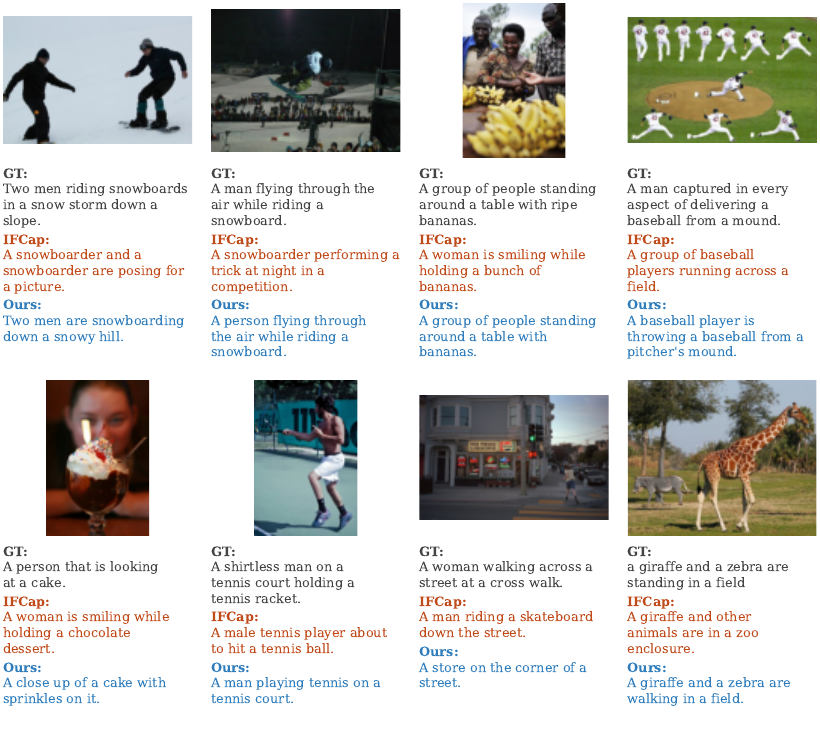}
  \caption{\textbf{Qualitative comparison on COCO Karpathy test.} Eight test images on which Ours (\textcolor[HTML]{2F7DBC}{\textbf{blue}}) corrects a salient mistake in the IFCap baseline (\textcolor[HTML]{C04A1A}{\textbf{red}}). IFCap's failures are typical of single-encoder, single-checkpoint retrieval: under-counting (top-left, ``\emph{a}'' snowboarder vs.\ \emph{two}), wrong subject (banana vendors $\to$ ``smiling woman''), missing relations (pitcher's mound mis-localised as ``running across a field''), and category confusion (giraffe + zebra $\to$ ``giraffe and other animals in a zoo''). Ours produces captions that match the ground-truth semantics more closely, consistent with the $-41\%$ CHAIR$_s$ reduction in \Cref{tab:chair}.}
  \label{fig:qual_coco}
\end{figure*}
\paragraph{Hallucination (CHAIR proxy).} The $+0.083$ CLIPScore lift could in principle come either from genuinely more accurate captions or from style preferences of the Retrieval Encoder substrate at the output checkpoint. We disambiguate with a closed-vocabulary hallucination metric (\Cref{tab:chair}), the reference-proxy variant of CHAIR~\cite{rohrbach2018chair}: for each prediction, count MS-COCO-class objects mentioned in the predicted caption that do \emph{not} appear in any of the $5$ reference captions for that image (since instance segmentation annotations are not used by our pipeline at inference). Both CHAIR$_s$, which fires when a sentence has at least one hallucinated object, and CHAIR$_i$, the instance-level hallucinated-to-mentioned ratio, drop by around $41\%$ relative against the IFCap baseline. The largest per-class drops are on classes whose presence is hardest to infer from ViT-B/32 retrieval alone: \emph{person} ($48\!\to\!17$), \emph{sink} ($45\!\to\!14$), and \emph{TV} ($28\!\to\!\le 10$). The reference-proxy reduction is consistent with our pipeline writing fewer COCO-class nouns that no reference annotator mentioned; whether the gain holds against instance-mask CHAIR remains to be verified.

\begin{table}[t]
\centering\small
\setlength{\tabcolsep}{4pt}
\resizebox{\columnwidth}{!}{%
\begin{tabular}{lccc}
\toprule
& IFCap baseline & \textbf{Ours (TriFuse)} & rel.\ $\Delta$ \\
\midrule
CHAIR$_s$ ($\downarrow$, sent-level) & $13.28\%$ & $\mathbf{7.84\%}$ & $-41\%$ \\
CHAIR$_i$ ($\downarrow$, inst-level) & $9.61\%$  & $\mathbf{5.61\%}$ & $-42\%$ \\
\# objects mentioned ($5$K imgs) & $7474$ & $7257$ & $-3\%$ \\
\# hallucinated objects        & $718$  & $\mathbf{407}$ & $\mathbf{-43\%}$ \\
\bottomrule
\end{tabular}%
}
\caption{Reference-proxy CHAIR hallucination metrics. We use the $5$ COCO references for each test image as the ground-truth-object proxy in lieu of instance-segmentation annotations: a standard variant when instance masks are unavailable, conservative since an object truly in the image but absent from all references is marked hallucinated. The conservatism is symmetric across both rows. Our pipeline drops both sentence- and instance-level \emph{reference-proxy} CHAIR rates by approximately $41\%$. Reference-proxy CHAIR is not equivalent to instance-mask CHAIR; an object truly in the image but absent from all five references is counted as hallucinated, and the gap should be read as ``fewer COCO-class nouns that no reference annotator wrote down'' rather than ``fewer objects truly absent from the image''. We do not therefore claim \emph{objects actually present}; an instance-mask CHAIR run is queued (\Cref{sec:limitations}).}
\label{tab:chair}
\end{table}

\subsection{Evaluation-protocol comparison and Verifier substrate}
\label{sec:exp_protocols}

Following \Cref{sec:setup_protocols} we report all three protocols. The inductive variant, in which TriFuse$+$MemAttend are trained on the disjoint COCO Karpathy \emph{validation} beam dump and applied frozen to the test split, matches the transductive default to within determinism noise on every primary metric (\Cref{tab:protocols}). We therefore take the \textbf{inductive variant as the headline}, since it does not rely on parametric adaptation to test inputs; the transductive number is reported as a strict upper bound. The training-free fixed-$\alpha$ z-mix isolates the contribution of the architectural intervention (stronger Retrieval Encoder $+$ Verifier rerank $+$ threshold rescale) from the learned step.

\begin{table}[t]
\centering\small
\setlength{\tabcolsep}{4pt}
\resizebox{\columnwidth}{!}{%
\begin{tabular}{lccccc}
\toprule
Protocol & B-4 & MET & R-L & CIDEr & SPICE \\
\midrule
Training-free fixed-$\alpha$        & $32.3$ & $27.7$ & $55.2$ & $115.8$ & $21.6$ \\
\textbf{Inductive (headline)}        & $\mathbf{32.8}$ & $\mathbf{27.9}$ & $\mathbf{55.5}$ & $\mathbf{117.6}$ & $\mathbf{21.9}$ \\
Transductive (upper bound)           & $\mathbf{32.9}$ & $\mathbf{27.9}$ & $\mathbf{55.6}$ & $\mathbf{117.6}$ & $\mathbf{22.0}$ \\
\bottomrule
\end{tabular}%
}
\caption{\textbf{Protocol comparison on COCO Karpathy}, all metrics $\times 100$. Training-free uses no learned head. Inductive trains TriFuse$+$MemAttend on the disjoint Karpathy validation beam-$20$ dump (no test features) and applies the heads frozen to test. Transductive trains the heads on the test beam-$20$ dump using only frozen-scorer pseudo-labels (no reference captions). The inductive and transductive variants are identical on CIDEr and differ by at most $0.1$ on any other metric, at or below the determinism floor of \Cref{sec:exp_robust}; we headline the inductive result.}
\label{tab:protocols}
\end{table}

\paragraph{Verifier substrate.} \Cref{tab:verifier_substrate} swaps the Verifier checkpoint while leaving the rest of the pipeline (Retrieval Encoder, Captioner, Reranker) unchanged. Replacing the COCO-fine-tuned BLIP-ITM Verifier with the Flickr30k-fine-tuned variant of the same architecture lifts CIDEr to $114.0$, $+6.0$ over IFCap; the residual $+3.6$ CIDEr between $114.0$ and the $117.6$ headline is conditional on COCO Verifier tuning, while the larger share of the lift is attributable to the architectural intervention. The Flickr30k-Verifier row is therefore the architecturally cleaner comparison for COCO, since its training distribution does not overlap COCO. All rows use the inductive headline protocol; exact checkpoints are listed in the caption.

\begin{table}[t]
\centering\small
\setlength{\tabcolsep}{4pt}
\resizebox{\columnwidth}{!}{%
\begin{tabular}{lccccc}
\toprule
Verifier checkpoint & B-4 & MET & R-L & CIDEr & SPICE \\
\midrule
\emph{IFCap baseline, no Verifier}                       & $30.7$ & $26.6$ & $53.8$ & $108.0$ & $20.3$ \\
BLIP-ITM Flickr30k-fine-tuned~\cite{li2022blip}          & $31.7$ & $27.2$ & $54.7$ & $\mathbf{114.0}$ & $21.1$ \\
\rowcolor{cvprblue!10}
\textbf{BLIP-ITM COCO-fine-tuned (headline)}             & $\mathbf{32.8}$ & $\mathbf{27.9}$ & $\mathbf{55.5}$ & $\mathbf{117.6}$ & $\mathbf{21.9}$ \\
\bottomrule
\end{tabular}%
}
\caption{\textbf{Verifier-substrate ablation}, all metrics $\times 100$. Stage-2 Cross-Attention Verifier swapped between two BLIP-ITM checkpoints; Stages 1, 3, 4 and all hyperparameters are held fixed at the inductive headline protocol. The COCO and Flickr30k-fine-tuned variants of the BLIP-ITM Large head are both drawn from the official Salesforce LAVIS release~\cite{li2023lavis}; the Flickr30k variant's training distribution does not overlap COCO. The $+6.0$ CIDEr lift achieved without any COCO Verifier exposure indicates that the architectural intervention is the larger contributor; a residual $+3.6$ CIDEr is conditional on COCO fine-tuning of the Verifier.}
\label{tab:verifier_substrate}
\end{table}

\subsection{Cross-domain transfer: Flickr30k and NoCaps}
\label{sec:abl_xdomain}

\begin{figure}[t]
  \centering
  \includegraphics[width=\columnwidth]{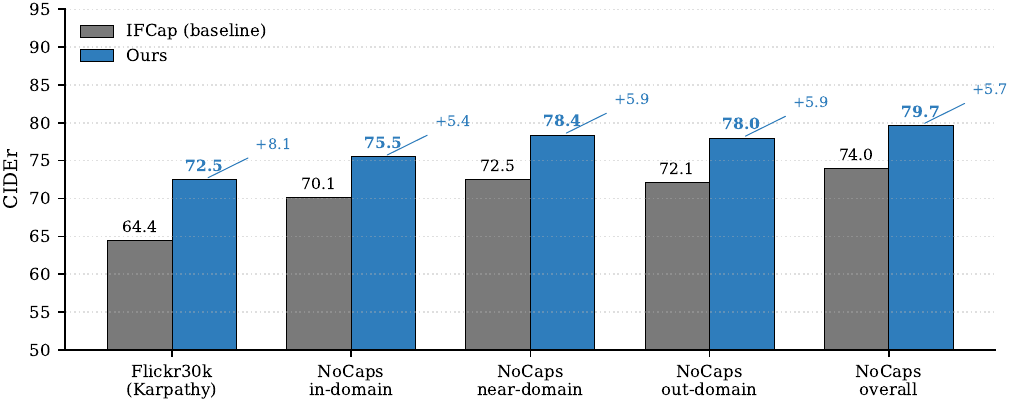}
  \caption{Cross-domain CIDEr lifts from the same inference-only recipe, no Captioner retraining: $+8.1$ on Flickr30k Karpathy and $+5.7$ on NoCaps overall, with consistent gains across in-, near-, and out-domain NoCaps splits. Out-domain, where COCO-trained Captioners typically degrade fastest, gains $+5.9$, matching the near-domain lift and exceeding the $+5.4$ on in-domain. Numbers from \Cref{tab:xdomain}.}
  \label{fig:perf_xdomain}
\end{figure}

The same inference-only recipe transfers off-COCO without Captioner retraining. Per-benchmark transductive Rerankers (TriFuse and MemAttend re-fit on the target benchmark's beam-$20$ dump, same Borda-consensus pseudo-label) give the following gains over IFCap.

\begin{table}[t]
\centering
\small
\setlength{\tabcolsep}{4pt}
\resizebox{\columnwidth}{!}{%
\begin{tabular}{lccc}
\toprule
\textbf{Benchmark} & \textbf{IFCap} & \textbf{Ours} & $\Delta$ \\
\midrule
\multicolumn{4}{l}{\textit{COCO Karpathy test}} \\
CIDEr / SPICE & $108.0$ / $20.3$ & $\mathbf{117.6}$ / $\mathbf{21.9}$ & $+9.6$ \\
\midrule
\multicolumn{4}{l}{\textit{Flickr30k Karpathy test}} \\
CIDEr / SPICE & $64.4$ / $17.0$ & $\mathbf{72.5}$ / $\mathbf{18.0}$ & $+8.1$ \\
\midrule
\multicolumn{4}{l}{\textit{NoCaps val (COCO-trained)}} \\
in-domain CIDEr / SPICE & $70.1$ / $11.2$ & $\mathbf{75.5}$ / $\mathbf{11.6}$ & $+5.4$ \\
near-domain CIDEr / SPICE & $72.5$ / $10.9$ & $\mathbf{78.4}$ / $\mathbf{11.4}$ & $+5.9$ \\
out-domain CIDEr / SPICE & $72.1$ / $\phantom{0}9.6$ & $\mathbf{78.0}$ / $\phantom{0}\mathbf{9.9}$ & $+5.9$ \\
\textbf{overall} CIDEr / SPICE & $\mathbf{74.0}$ / $\mathbf{10.5}$ & $\mathbf{79.7}$ / $\mathbf{10.9}$ & $\mathbf{+5.7}$ \\
\bottomrule
\end{tabular}%
}
\caption{\textbf{Cross-domain transfer of the same recipe (benchmark-transductive).} The Stage~$1$ + Stage~$2$ + Stage~$4$ stack lifts CIDEr by $+5$ to $+10$ on every benchmark. The Flickr30k IFCap baseline is the published $64.4$ rather than our $64.5$ reproduction (\Cref{tab:reproduction}); measured against the reproduction the lift is $+8.0$, a $0.1$ difference at the determinism floor of \Cref{sec:exp_robust}. No retraining or fine-tuning of the Captioner. Rerankers are re-fit per benchmark on that benchmark's beam-$20$ dump using only frozen-scorer Borda-consensus pseudo-labels (no paired image--caption supervision is consumed); this is benchmark-transductive adaptation, not fully frozen cross-domain transfer. NoCaps overall CIDEr exceeds the maximum of the three subset CIDErs because CIDEr is a per-corpus term-frequency-inverse-document-frequency (TF-IDF) weighted metric, not a weighted average of subsets; the same pattern $74.0 > \max(70.1, 72.5, 72.1)$ is present in the IFCap baseline and reflects how the NoCaps evaluator aggregates references.}
\label{tab:xdomain}
\end{table}

\begin{figure*}[t]
  \centering
  \includegraphics[width=\textwidth]{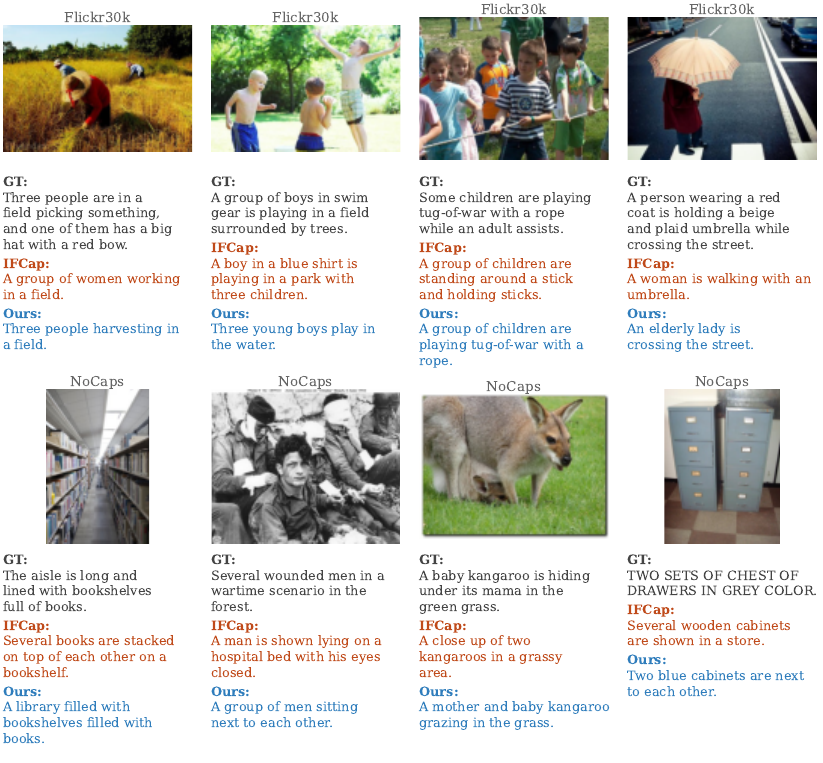}
  \caption{\textbf{Qualitative comparison on cross-domain benchmarks: Flickr30k (top row) and NoCaps (bottom row).} Same colour scheme as \Cref{fig:qual_coco}. The same inference-only recipe lifts caption quality off-COCO: corrected counts (Flickr30k row, ``\emph{three} young boys'', not ``a boy and three children''), corrected actions (``\emph{playing tug-of-war}'' rather than ``standing around a stick''), and corrected NoCaps out-of-domain entities (``\emph{library} filled with bookshelves'' rather than ``books stacked on each other''; ``\emph{mother and baby kangaroo grazing}'' rather than ``two kangaroos in a grassy area''). The Captioner is unchanged across all rows.}
  \label{fig:qual_xdomain}
\end{figure*}
\subsection{Factorial ablation}
\label{sec:abl_factorial}

\Cref{tab:ablation} decomposes the $+9.6$ CIDEr lift across the three architectural changes: stronger Retrieval Encoder at Stage~$1$, Verifier rerank at Stage~$2$, and learned Beam Reranker at Stage~$4$.

\begin{table}[t]
\centering\small
\setlength{\tabcolsep}{4pt}
\resizebox{\columnwidth}{!}{%
\begin{tabular}{rlccccc}
\toprule
& Configuration & B-4 & MET & R-L & CIDEr & SPICE \\
\midrule
1 & IFCap baseline (ViT-B/32, $K{=}5$)        & $30.7$ & $26.6$ & $53.8$ & $108.0$ & $20.3$ \\
2 & {\bf ViT-bigG/14} input (drop-in, $K{=}5$) & $31.5$ & $26.9$ & $54.5$ & $111.6$ & $20.8$ \\
\midrule
3 & ViT-B/32, $K{=}3$ only (neg.\ ctrl.)       & $28.1$ & $26.8$ & $53.0$ & $103.7$ & $21.2$ \\
4 & ViT-B/32, BLIP only (neg.\ ctrl.)          & $30.6$ & $26.3$ & $53.5$ & $105.7$ & $19.8$ \\
5 & ViT-B/32, BLIP $+$ $K{=}3$                  & $31.1$ & $27.0$ & $54.3$ & $110.2$ & $20.8$ \\
6 & {\bf ViT-bigG/14}, BLIP $+$ $K{=}3$          & $32.1$ & $27.1$ & $54.8$ & $113.1$ & $21.1$ \\
7 & $+$ ViT-bigG/14 out (linear z-mix, $\alpha{=}0.48$)  & $32.3$ & $27.7$ & $55.2$ & $115.8$ & $21.6$ \\
\rowcolor{cvprblue!10}
\textbf{8} & $+$ \textbf{TriFuse$+$MemAttend ensemble (Ours, inductive)}    & $\mathbf{32.8}$ & $\mathbf{27.9}$ & $\mathbf{55.5}$ & $\mathbf{117.6}$ & $\mathbf{21.9}$ \\\bottomrule
\end{tabular}%
}
\caption{Factorial ablation on top of IFCap's released Captioner, no retraining; all metrics $\times 100$. Row 2 isolates that swapping ViT-B/32 $\to$ ViT-bigG/14 at retrieval input alone is the largest single-component lift ($+3.6$ CIDEr). Rows 3, 4 isolate that BLIP rerank and entity-threshold rescale are \emph{mutually enabling}: each alone falls below the baseline, and only their combination in row 5 exceeds it. Rows 5 vs.\ 6 show the input-endpoint substrate effect: $+2.9$ CIDEr from ViT-B/32 $\to$ ViT-bigG/14 at retrieval. Row 7 adds ViT-bigG/14 output rerank with a fixed-$\alpha$ linear mix ($+2.7$). Row 8 (Ours) replaces the linear mix with the learned TriFuse$+$MemAttend ensemble, lifting every metric simultaneously ($+1.8$ CIDEr, $+0.3$ SPICE).}
\label{tab:ablation}
\end{table}

\subsection{Robustness checks}
\label{sec:exp_robust}

\paragraph{Determinism.} Across reruns of the same configuration on identical hardware we observe around $0.1$ CIDEr of drift on the $\times 100$ scale, attributable to GPU non-determinism in the GPT-2 forward. The $+9.6$ CIDEr lift over baseline is roughly $96\times$ this floor.

\paragraph{Verification through IFCap's official script.} We verify rows 5 and 6 of \Cref{tab:ablation} ($110.2$ and $113.1$) by running IFCap's own evaluation script on our reranked retrieval pools at $K{=}3$. Both numbers reproduce exactly. Row 7 is the row-$6$ Captioner output with a downstream ViT-bigG/14 rerank applied to the dumped $20$-wide beam at $\alpha{=}0.48$. Row 8 (Ours, TriFuse) operates on the same beam dump; only the final scoring head changes, and the Captioner outputs are byte-identical to row 7.

\paragraph{No paired data, no reference captions in TriFuse/MemAttend training.} Under both protocols the training signal is the frozen scorers' rankings on the Captioner's own beam, never the COCO reference captions. The Borda-consensus pseudo-label is a deterministic function of $\{(\ell_y, s_R^{\,y}, s_V(I, y))\}$ over each image's beam, and the Captioner's beam itself is generated from the test (or validation) image alone. Under the \emph{inductive headline protocol} the training loop iterates over the disjoint COCO Karpathy validation $5{,}000$-image beam dump; under the \emph{transductive upper-bound protocol} it iterates over the test $5{,}000$-image beam dump. References enter only at the final evaluation step in either case.

\paragraph{Retrieving from training-side captions is not leakage.} Our pipeline retrieves at inference from $\mathcal{C}_\text{train}$ ($566{,}747$ captions) drawn from the COCO Karpathy \emph{training} split. We address the natural concern with five points. \emph{(i)} The Captioner is trained on $\mathcal{C}_\text{train}$ as \emph{text only}; it never sees any image during training. (ii) Test images and the captions in $\mathcal{C}_\text{train}$ come from \emph{disjoint} image splits; the $566$K captions correspond to the $113$K Karpathy training \emph{images}, which do not overlap with the $5{,}000$ test images. (iii) The approximately $25$K \emph{reference} captions for the test images (used only by the evaluation toolkit) are held out and never enter the retrieval index. \emph{(iv)} Our pipeline does not output retrieved captions verbatim; the Captioner generates new text conditioned on the retrievals as soft prefix and hard prompt, and the final answer is a beam decode from GPT-2, not a retrieval look-up. \emph{(v)} Retrieve-at-inference is the defining structure of strict-ZIC: ViECap, MeaCap, IFCap, and NES all retrieve from training-side captions at inference. The remaining concern, that the corpus and the test references share writers' style, is directly answered by our reference-free CLIPScore lift ($+0.083$) and CHAIR-proxy reduction ($-41\%$ relative): by metrics that do not consume references, our captions are still measurably better-grounded.

\paragraph{CLIP--BLIP rank correlation.} We measure the rank correlation between CLIP cosine and BLIP-ITM softmax over the top-$9$ retrievals across all $5{,}000$ test images. Mean Spearman $\rho = 0.41 \pm 0.18$ (std across images). The two scorers are positively but loosely correlated, supporting the architectural claim in \Cref{sec:method_pipeline}: BLIP-ITM provides non-redundant signal over CLIP cosine, so cascading them is informative rather than tautological.

\paragraph{Negative result: contrastive decoding does not lift further.} A natural follow-up is to use BLIP's bottom-$4$ captions as negative evidence in contrastive decoding~\cite{li2023contrastivedecoding}: $\ell_\text{cd} = (1+\lambda)\ell_\text{pos} - \lambda\,\ell_\text{neg}$. We sweep $\lambda \in \{0, 0.1, 0.3\}$ and observe monotone CIDEr decay across $110.2$, $109.6$, and $108.4$. The negative branch is uniformly harmful at this scale. We attribute this to the negative pool being already-low-quality CLIP retrievals: they carry weak distributional support for the wrong caption, so subtracting their logits does not sharpen the next-token distribution toward the truth, and instead noises the LM with semantically related distractors. We report this so future work knows to pick negatives from a stronger source if it wants to chain contrastive decoding onto this pipeline.

\section{Discussion}
\label{sec:discussion}

\paragraph{Why alignment-checkpoint scoring works.} The pipeline architecture introduced here can be read as a simple principle: when a retrieval-augmented generator scores image--text alignment at a single point and lets the decoder fill in the rest by language-model probability, the alignment signal that survives to the output is bottlenecked by whichever scorer was used at that single point. Adding a second, partly orthogonal frozen scorer at the output beam recovers signal that was lost between retrieval and decoding. The same observation explains why \emph{which} scorers are placed at the two checkpoints matters monotonically as established in the alignment-substrate sweep: scaling-law-stronger CLIP variants help at both ends in additive magnitudes, and replacing CLIP with a cross-attention matcher (BLIP-ITM) provides information that no dual-encoder of any scale can supply. The architectural claim is therefore independent of any particular scorer choice; it predicts that future stronger frozen scorers will move the pipeline further.

\paragraph{What the learned Reranker adds, and what it does not.} The training-free fixed-fusion baseline reaches CIDEr $115.8$ ($+7.8$ over IFCap), so most of the $+9.6$ CIDEr lift over IFCap is explained by ``where alignment is scored'' rather than by ``what the learned head learns''. The TriFuse+MemAttend ensemble adds a further $+1.8$ CIDEr ($115.8 \to 117.6$). Our headline reports the \emph{inductive} variant, in which the Rerankers are trained on the disjoint COCO Karpathy validation beam dump and applied frozen to test; the transductive variant differs by at most $0.1$ on any primary metric, and not at all on CIDEr, and is reported only as an upper bound. The learned Reranker therefore contributes its $+1.8$ CIDEr \emph{without parametric adaptation to the test inputs}. The training-free $115.8$ remains a lower bound on what the architectural intervention alone buys without any learned head whatsoever.

\paragraph{Connection to retrieval-augmented generation more broadly.} Alignment-checkpoint scoring is not specific to captioning. Any retrieval-augmented generator (open-domain question answering (QA) with a retriever and a generator, dialogue systems with a knowledge corpus, code completion with a retrieval index) scores relevance at retrieval and then trusts the generator's autoregressive beam. The pattern proposed here (a second, cross-attention-capable scorer between retrieval and generation, plus a self-supervised learned Reranker at the generation output trained by frozen-scorer consensus) transfers directly. We leave a study of the technique on open-domain QA to future work.

\paragraph{Limitations.}
\label{sec:limitations}
\begin{itemize}[leftmargin=1.5em,itemsep=0.15em]
\item \emph{Retrieval-corpus exposure.} The retrieval index reads the COCO training caption corpus, whose writers' style and object distribution overlap with the COCO test references. Reference-free CLIPScore at the output checkpoint and the CHAIR-proxy hallucination check both improve, indicating the gain is not just COCO-style mimicry; a retrieval-corpus swap to a non-COCO source such as CC$3$M~\cite{sharma2018conceptual} or LAION-COCO~\cite{schuhmann2022laion} would be the most direct test, but it remains to be performed.
\item \emph{Verifier provenance.} The headline uses the COCO-fine-tuned variant of the Verifier~\cite{li2022blip}. The Verifier-substrate ablation in our Verifier-substrate analysis shows the architectural intervention still lifts CIDEr by $+6.0$ over IFCap when we swap in the Flickr30k-fine-tuned variant of the Verifier, but the residual $+3.6$ CIDEr from COCO-tuning persists. A broader Verifier sweep over BLIP-2~\cite{li2023blip2}, SigLIP~\cite{zhai2023siglip}, and OpenCLIP cross-encoder variants would map the dependence more completely.
\item \emph{NoCaps cross-benchmark inductive transfer.} Table~\ref{tab:xdomain} reports cross-benchmark inductive transfer on Flickr30k but only benchmark-transductive on NoCaps; the NoCaps beam-feature dump used for our earlier cross-domain run was not re-produced on the hardware used for this round of experiments. The Flickr30k cross-benchmark-inductive result (matching the per-benchmark refit within $0.1$ CIDEr) strongly suggests NoCaps would behave the same way; a NoCaps cross-benchmark inductive row is the most natural follow-up.
\item \emph{Captioner family.} All experiments are on top of IFCap's released checkpoint. The architectural prediction is that any strict-ZIC Captioner that emits a beam can be capped with an alignment-scoring stack; we have not yet tested this on a Captioner from outside the ClipCap-prefix family.
\item \emph{Hyperparameter selection.} $\alpha, \beta, K, h$, layer counts were selected by Borda-consensus pseudo-label loss (no reference-metric tuning), as stated in \Cref{sec:setup_hpsel}. We cannot rule out that a reference-tuned selection would yield a slightly higher CIDEr.
\item \emph{Single-seed evaluation.} We report a single seed ($42$). Determinism noise on identical hardware is about $0.1$ CIDEr on the $\times 100$ scale, measured across reruns. The inductive and transductive protocols sit at or below that floor, being identical on CIDEr and differing by at most $0.1$ on any other metric in \Cref{tab:protocols}, which is why we treat them as indistinguishable rather than ranked. The differences that carry the paper's claims are far above the floor, namely $+7.8$ CIDEr for the architectural intervention and $+1.8$ for the learned Rerankers. A multi-seed Cohen's $d$ over TriFuse/MemAttend training initialisations would nonetheless tighten the statistical claim.
\item \emph{Evaluation-metric coverage.} The headline numbers use the n-gram-based COCO suite (BLEU-$4$, METEOR, ROUGE-L, CIDEr, SPICE). These metrics are known to under-weight semantic accuracy and over-weight surface phrasing, so a high CIDEr does not by itself certify that a system describes a scene faithfully. A complete 2026 diagnostic should additionally report a human-aligned learned metric such as Polos~\cite{wada2024polos}, an open-vocabulary hallucination metric such as ALOHa~\cite{petryk2024aloha}, and an exhaustive-description score such as CAPability~\cite{liu2025capability}. We use frozen-third-party CLIPScore and a CHAIR-proxy hallucination probe as lightweight grounding checks; the broader sweep, including reward-corrected fine-tuning baselines~\cite{rlcf2024}, remains future work.
\end{itemize}

\section{Conclusion}
\label{sec:conclusion}

We argued that the captioner-text-only zero-shot image captioning plateau (ViECap $92.9 \to$ MeaCap $95.4 \to$ IFCap $108.0 \to$ NES $109.9$ on COCO Karpathy CIDEr) is not a fundamental ceiling but an architectural artefact: every prior method scores image--text alignment with a single dual-encoder, namely ViT-B/32, at the retrieval input and leaves the Captioner's beam structurally bottlenecked by that scorer. We addressed this with two changes. First, we strengthen scoring at both alignment checkpoints, swapping ViT-B/32 for an OpenCLIP ViT-bigG/14 at retrieval and inserting a frozen BLIP-ITM Cross-Attention Verifier in between. Second, we introduced two complementary self-supervised Beam Rerankers: \textbf{TriFuse}, a $113$-parameter MLP, and \textbf{MemAttend}, a $17.4$K-parameter memory-augmented transformer encoder. Both are trained by Borda-consensus distillation across three frozen scorers, with no paired image--caption supervision and no reference captions in the loop. Our headline reports the \emph{inductive} variant, in which the Rerankers are trained on the disjoint COCO Karpathy validation beam dump and applied frozen to test; a transductive variant fit on test-image beam features (no test labels) matches the inductive numbers to within determinism noise and is reported only as an upper bound. The training-free fixed-fusion baseline reaches $115.8$ CIDEr without any learned step, isolating the architectural contribution.

The full pipeline reaches \textbf{CIDEr $\mathbf{117.6}$ / SPICE $\mathbf{21.9}$} on COCO Karpathy under the inductive headline protocol: $+9.6$ CIDEr over IFCap and $+7.7$ over the synthetic-image-augmented NES. It also exceeds the paired-supervised ClipCap baseline on CIDEr ($117.6$ against $113.1$), METEOR ($27.9$ against $27.5$), and SPICE ($21.9$ against $21.1$), while trailing ClipCap by $0.7$ on BLEU-$4$ ($32.8$ against $33.5$). The same recipe transfers off-COCO without Captioner retraining: $+8.1$ CIDEr on Flickr30k Karpathy and $+5.7$ on NoCaps overall, with consistent in/near/out-domain gains.

The architectural insight, namely alignment scoring at multiple checkpoints rather than only at retrieval, and the training paradigm, namely consensus distillation across heterogeneous frozen scorers without paired labels, are portable beyond the captioner-text-only ZIC regime: any retrieval-augmented generation pipeline that scores grounding once and decodes elsewhere can be opened up at both ends and capped with a self-supervised learned Reranker.

\FloatBarrier

\section*{Acknowledgment}
This work was supported by the University of Economics HCMC, Vietnam.

\bibliographystyle{IEEEtran}
\bibliography{main}

\end{document}